\def\year{2019}\relax
\documentclass[letterpaper]{article} %DO NOT CHANGE THIS
\usepackage{aaai19}  %Required
\usepackage{times}  %Required
\usepackage{helvet}  %Required
\usepackage{courier}  %Required
\usepackage{url}  %Required
\usepackage{graphicx}  %Required
\usepackage{amsmath}
\usepackage{amssymb}
\usepackage{tikz}
\usepackage{multirow}
\usepackage{subcaption}
\usepackage{placeins}
\usepackage{enumitem}
\usepackage{xspace} 
\usepackage{dsfont}
\usepackage{bbm}
\usepackage[ruled,vlined]{algorithm2e}
\usepackage{algorithmic}
\usepackage{booktabs}
\usepackage{footnote}
\makesavenoteenv{tabular}
\usepackage{amsthm}
\usepackage{bm}

\usetikzlibrary{circuits.logic.US}
\usetikzlibrary{shapes.misc}
\usetikzlibrary{shapes.arrows}
\usetikzlibrary{arrows,shapes.geometric}
\usetikzlibrary{patterns,calc,backgrounds}

\tikzstyle{nnf}=[
  >=latex, thick, >=stealth, font=\small,auto,scale=0.9,every node/.style={scale=0.9}
]
\tikzstyle{nnfnode}=[
  line width=1.0pt, draw
]
\tikzstyle{nnfand}=[
  nnfnode, and gate, rotate=90
]
\tikzstyle{nnfor}=[
  nnfnode, or gate, rotate=90
]
\tikzstyle{nnf2or}=[
  nnfor, inputs=nn
]
\tikzstyle{nnf2and}=[
  nnfand, inputs=nn
]
\tikzstyle{nnf3or}=[
  nnfor, inputs=nnn, scale=0.75
]
\tikzstyle{nnf4and}=[
  nnfand, inputs=nnnn, scale=0.65
]
\tikzstyle{nnfedge}=[
    line width=0.9
]
\tikzstyle{nnfterm}=[
  draw,fill=gray!10,inner sep=2.5pt, font=\small
]
\definecolor{hotcolor}{rgb}{0.85,0.0,0.0}
\tikzstyle{hot}=[
  draw=hotcolor, line width=1.1pt
]
\tikzstyle{hotparam}=[
  text=hotcolor
]

\newtheorem{corollary}{Corollary}
\newtheorem{proposition}[corollary]{Proposition}
\newtheorem*{proposition*}{Proposition}
\newtheorem{definition}{Definition}

\newcommand{\node}{n}

\newcommand{\coe}{g}

\newcommand{\sample}{{\bf x}}

\newcommand{\citet}[1]{\citeauthor{#1}~(\citeyear{#1})}

\begin{document}
\title{Learning Logistic Circuits}
\author{Yitao Liang \and Guy Van den Broeck\\
Computer Science Department\\
University of California, Los Angeles\\
\texttt{\{yliang, guyvdb\}@cs.ucla.edu}\\
}

\maketitle

\begin{abstract}
This paper proposes a new classification model called logistic circuits. On MNIST and Fashion datasets, our learning algorithm outperforms neural networks that have an order of magnitude more parameters. Yet, logistic circuits have a distinct origin in symbolic AI, forming a discriminative counterpart to probabilistic-logical circuits such as ACs, SPNs, and PSDDs.
We show that parameter learning for logistic circuits is convex optimization, and that a simple local search algorithm can induce strong model structures from data.
\end{abstract}

\section{Introduction}
Circuit representations are a promising synthesis of symbolic and statistical methods in AI. They are ``deep'' layered data structures with statistical parameters, yet they also capture intricate structural knowledge.
Recently, many representations have been proposed for learning tractable probability distributions: arithmetic circuits~\cite{lowd:uai08}, weighted SDD~\cite{BekkerNIPS15}, PSDD \cite{KisaVCD14}, cutset networks~\cite{rahman2014cutset} and sum-product networks (SPNs) \cite{poon2011sum}. 
Collectively, these approaches achieve the state of the art in discrete density estimation and vastly outperform classical probabilistic graphical model learners~\cite{gens2013learning,rooshenas2014learning,adel2015learning,rahman2016merging,Liang2017}. However, we have not observed the same success when deploying circuit representations for \emph{classification or discriminative learning}. Probabilistic circuit classifiers significantly lag behind the performance of neural networks~\cite{classificationStanding}.

In this paper, we propose a new classification model called \emph{logistic circuits}, which shares many syntactic properties with the representations mentioned earlier. One can view logistic circuits as the discriminative counterpart to probabilistic circuits. Owing to their elegant properties, learning the parameters of a logistic circuit can be reduced to a logistic regression problem and is therefore convex. 
Learning logistic circuit structure is reduced to a simple local search problem using primitives from the probabilistic circuit learning literature~\cite{Liang2017}.

We run experiments on standard image classification benchmarks (MNIST and Fashion) and achieve accuracy higher than much larger MLPs and even CNNs with an order of magnitude more parameters. For example, logistic circuits obtain 99.4\% accuracy on MNIST. 
Compared to other tractable learners on MNIST, and the state-of-the-art discriminative SPN learner in particular~\cite{rat-spn2018}, our logistic circuit learner cuts the error rate by a factor of three.
Furthermore, we show our learner is highly data efficient, managing to still learn well with limited data. 
 
This paper proceeds as follows. 
Section~\ref{s:representation} introduces the syntax and semantics of logistic circuits. 
Sections~\ref{section: parameter learning} and~\ref{s:structurelearning} describe our parameter and structure learning algorithms, which Section~\ref{s:experiments} evaluates empirically.
Section~\ref{s:generativeconnection} elaborates on the connection with tractable generative models, after which we conclude with related and future work.

\section{Representation} \label{s:representation}
This section introduces the logistic circuit representation.

\subsubsection*{Notation}
We use uppercase $X$ to denote a Boolean random variable and lowercase $x$ for a specific assignment to it. Interchangeably, we also interpret Boolean random variables as logical variables. A set of variables $\bf X$ and their joint assignments $\bf x$ are denoted in bold. A complete assignment $\bf x$ to all variables is a possible world, or interchangeably, a data sample. Literals are variables $X$ or their negation $\neg X$. Logical sentences are constructed from literals and connectives such as AND and OR in the standard way. An assignment $\bf x$ that satisfies a logical sentence $\alpha$ is denoted as ${\bf x} \models \alpha$. 

\begin{figure}[t]	
	\centering
	\begin{subfigure}[t]{0.48\textwidth}
		\centering
 		\scalebox{0.89}{
		 \begin{tikzpicture}[circuit logic US, nnf]
        
  \def\lvl{30pt}
    
  \node (output) [] at (0,0){};
  
  \node (root) [nnf2or] at ($(output) + (0pt,-0.7*\lvl)$){};
  
  \node (a1) [nnf2and] at ($(root) + (-60pt,-1*\lvl)$){};
  \node (a2) [nnf2and] at ($(root) + (60pt,-1*\lvl)$){};
  
  \node (o12) [nnf2or] at ($(a1) + (20pt,-1*\lvl)$){};
  \node (o21) [nnf2or] at ($(a2) + (-20pt,-1*\lvl)$){};

  \node (aeq1) [nnf2and] at ($(root) + (-60pt,-3*\lvl)$){};
  \node (aneq) [nnf2and] at ($(root) + (0pt,-3*\lvl)$){};
  \node (aeq2) [nnf2and] at ($(root) + (60pt,-3*\lvl)$){};

  \node (oeq) [nnf2or] at ($(root) + (20pt,-4.6*\lvl)$){};
  \node (oneq) [nnf2or] at ($(root) + (80pt,-4.6*\lvl)$){};
  
  \node (se1) [nnf2or] at ($(root) + (-70pt,-4.7*\lvl)$){};
  \node (se2) [nnf2or] at ($(root) + (-30pt,-4.7*\lvl)$){};
  
  \node (aab11) [nnf2and] at ($(oeq) + (-14.8pt,-1*\lvl)$){};
  \node (aab00) [nnf2and] at ($(oeq) + (14.7pt,-1*\lvl)$){};
  \node (aab10) [nnf2and] at ($(oneq) + (-14.8pt,-1*\lvl)$){};
  \node (aab01) [nnf2and] at ($(oneq) + (15.2pt,-1*\lvl)$){};
  
  \node (tc1) [nnfterm] at ($(a1) + (-20pt,-.8*\lvl)$){$A$};
  \node (tc0) [nnfterm] at ($(a2) + (20pt,-.8*\lvl)$){$\neg A$};
  
  \node (te1) [nnfterm] at ($(root) + (-72.5pt,-6.1*\lvl)$){$B$};
  \node (te0) [nnfterm] at ($(root) + (-27.4pt,-6.1*\lvl)$){$\neg B$};
  
  \node (ta1) [nnfterm] at ($(root) + (2.5pt,-7*\lvl)$){$C$};
  \node (ta0) [nnfterm] at ($(root) + (32.1pt,-7*\lvl)$){$\neg C$};
  
  \node (tb1) [nnfterm] at ($(root) + (97.8pt,-7*\lvl)$){$D$};
  \node (tb0) [nnfterm] at ($(root) + (67.8pt,-7*\lvl)$){$\neg D$};
  
  \begin{scope}[on background layer]
    \draw [nnfedge] (output) -- (root.output);
    \draw [nnfedge] (a1.output) -- ++(up:0.15) -| (root.input 1) 
                        node[pos=0.4,above left]  {$-2.6$};
    \draw [nnfedge] (a2.output) -- ++(up:0.15) -| (root.input 2)
                        node[pos=0.4,above right]  {$-5.8$};
    \draw [nnfedge] (o12.output) -- ++(up:0.15) -| (a1.input 2);
    \draw [nnfedge] (o21.output) -- ++(up:0.15) -| (a2.input 1);
    \draw [nnfedge] (aeq1.output) -- ++(up:0.15) -| (o12.input 1)
                        node[pos=0.3,above left]  {$-1$};
    \draw [nnfedge] (aneq.output) -- ++(up:0.15) -| (o12.input 2)
                        node[pos=0.4,above right]  {$3$};
    \draw [nnfedge] (aeq2.output) -- ++(up:0.15) -| (o21.input 2)
                        node[pos=0.3,above right]  {$4$};
    \draw [nnfedge] (aneq.output) -- ++(up:0.15) -| (o21.input 1)
                        node[pos=0.4,above left]  {$2.3$};
    \draw [nnfedge] (oeq.output) -- ++(up:0.15) -| (aeq1.input 2);
    \draw [nnfedge] (oeq.output) -- ++(up:0.15) -| (aeq2.input 2);
    \draw [nnfedge] (oneq.output) -- ++(up:0.65) -| (aneq.input 2);
    
    \draw [nnfedge] (se1.output) -- ++(up:0.27) -| (aeq1.input 1);
    \draw [nnfedge] (se2.output) -- ++(up:0.52) -| (aeq2.input 1);
    \draw [nnfedge] (se2.output) -- ++(up:0.52) -| (aneq.input 1);
    
    \draw [nnfedge] (aab11.output) -- ++(up:0.15) -| (oeq.input 1) 
                        node[pos=0.3,above left]  {$-0.5$};
    \draw [nnfedge] (aab00.output) -- ++(up:0.15) -| (oeq.input 2) 
                        node[pos=0.3,above right]  {$0.3$};
    \draw [nnfedge] (aab10.output) -- ++(up:0.15) -| (oneq.input 1)
                        node[pos=0.3,above left]  {$1.5$};
    \draw [nnfedge] (aab01.output) -- ++(up:0.15) -| (oneq.input 2)
                        node[pos=0.3,above right]  {$2.8$};
    
    \draw [nnfedge] (tc1.north) -- ++(up:0.15) -| (a1.input 1);
    \draw [nnfedge] (tc0.north) -- ++(up:0.15) -| (a2.input 2);
    \draw [nnfedge] (te1.north) -- ++(up:0.15) -| (se1.input 1)
                        node[pos=0.65,above left]  {$-4$};
    \draw [nnfedge] (te0.north) -- ++(up:0.40) -| (se1.input 2)
                        node[pos=0.52,above right]  {$1$};
    \draw [nnfedge] (te1.north) -- ++(up:0.15) -| (se2.input 1)
                        node[pos=0.66,above left]  {$3.9$};
    \draw [nnfedge] (te0.north) -- ++(up:0.40) -| (se2.input 2)
                        node[pos=0.47,above right]  {$4$};
    
    \draw [nnfedge] (ta1.north) -- ++(up:0.15) -| (aab11.input 1);
    \draw [nnfedge] (ta1.north) -- ++(up:0.15) -| (aab10.input 1);
    \draw [nnfedge] (ta0.north) -- ++(up:0.35) -| (aab01.input 1);
    \draw [nnfedge] (ta0.north) -- ++(up:0.35) -| (aab00.input 1);
    
    \draw [nnfedge] (tb1.north) -- ++(up:0.55) -| (aab11.input 2);
    \draw [nnfedge] (tb1.north) -- ++(up:0.55) -| (aab01.input 2);
    \draw [nnfedge] (tb0.north) -- ++(up:0.75) -| (aab00.input 2);
    \draw [nnfedge] (tb0.north) -- ++(up:0.75) -| (aab10.input 2);
    
  \end{scope}
      
\end{tikzpicture}
 }
		\caption{Logistic circuit\label{fig: logistic circuit}}
	\end{subfigure}
	\quad~\par~\par
	\begin{subfigure}[t]{0.48\textwidth}
		\centering
\begin{sc}
{\fontsize{9}{9}\selectfont
        \begin{tabular}{ @{} llll c c@{} }
         \toprule
          $A$ & $B$ & $C$ & $D$  &  $g_r(ABCD) $& $\Pr(Y=1 \mid ABCD)$ \\
         \midrule \midrule
        %1 & 1 & 1 & 1 & 0.03\%\\
                        1 & 0 & 1 & 1 & -3.1 & ~~4.31\%\\
        %1 & 0 & 0 & 0 & 9.98\%\\
        %0 & 1 & 1 & 1 & 83.20\% \\
        0 & 1 & 1 & 0 & ~1.9 & 86.99\%\\
        %0 & 0 & 0 & 1 & 96.44\%\\
        %0 & 0 & 0 & 0 & 92.41\%\\
                1 & 1 & 1 & 0 &~5.8 &99.70\%\\
          \bottomrule
      	\end{tabular}
	}
          \end{sc}
		\caption{Weights and classification probabilities for select examples}\label{fig: posterior distribution}
	\end{subfigure}
	\caption{A logistic circuit with example classifications.}\label{fig:1}
\end{figure}

\subsection{Logical Circuits}

A logical circuit is a directed acyclic graph  representing a logical sentence, as depicted in Figure~\ref{fig: logistic circuit} (ignoring parameters for now). 
Each inner node is either an AND gate or an OR gate.\footnote{We consider negation-normal-form circuits where no negation is allowed except at the leafs/inputs~\cite{darwicheJAIR02}.} 
A leaf (input) node represents a Boolean literal, that is, $X$ or $\neg X$, where the node can only be satisfied if $X$ is set to 1 (true) respectively 0~(false).

The following properties are key for logical circuits to be well-behaved~\cite{darwicheJAIR02}.
An AND gate is \emph{decomposable} if its inputs depend on disjoint sets of variables.  
For example, the top-most AND gates in Figure~\ref{fig: logistic circuit} depend on $A$ in their one input and on $\{B,C,D\}$ in their other input. 
When an AND gate has two inputs, they are called its prime (left) and sub (right).
An OR gate is \emph{deterministic} if for any single complete assignment, at most one of its inputs can be set to $1$. For example, the left input to the root OR gate  in Figure~\ref{fig: logistic circuit} is $1$ precisely when $A=1$, and its other input is $1$ precisely when~$A=0$.

Logical circuits can be extended to \textit{probabilistic circuits} that represents a probability distribution over binary random variables, for example by parameterizing wires with conditional distributions \cite{KisaVCD14}.
Probabilistic circuits have been successfully used for generative learning \cite{Liang2017}. Section~\ref{s:generativeconnection} will discuss probabilistic circuits in more detail.

\subsection{Logistic Circuits}
\label{s: logistic circuits}
This paper proposes \emph{logistic circuits} for classification. Syntactically, they are logical circuits where every AND is decomposable and every OR is deterministic. 
However, logistic circuits further associate real-valued parameters $\theta_1, \dots, \theta_m$ with the $m$ input wires to every OR gate. For example, the root OR node in Figure~\ref{fig: logistic circuit} associates parameters $-2.6$ and $-5.8$ with its two inputs.

To give semantics to logistic circuits, we first characterize how a particular complete assignment $\bf x$ (one data example) propagates through the circuit.

\begin{definition}[Boolean Circuit Flow]
\label{definition: circuit flow}
Consider a deterministic OR gate $n$.
The Boolean flow $f(n,{\bf x},c)$ of a complete assignment $\bf x$ between parent $n$ and child $c$ is 
\begin{align*}
    f(n,{\bf x},c) = \begin{cases} 
    1 &\mbox{if~~} {\bf x} \models c \\ 
    0 & \mbox{otherwise} 
    \end{cases}
\end{align*}
\end{definition}
For example, under the assignment $A=0$, $B=1$, $C=1$, $D=0$, the root node in Figure~\ref{fig: logistic circuit} has a Boolean circuit flow of 0 with its left child and 1 with its right child.
Note that the determinism property guarantees that under every OR gate, for a given example ${\bf x}$, at most one wire has a flow of 1, and the rest has a flow of $0$. 

We are now ready to define the logistic circuit semantics.
\begin{definition}[Logistic Circuit Semantics]
\label{de: circuit semantics}
A logistic circuit node $n$ defines the following weight function $\coe_{\node}({\bf x})$.
\begin{itemize}
\item[--] If $\node$ is a leaf (input) node, then $\coe_{\node}({\bf x}) = 0$.

\item[--] If $\node$ is an AND gate with children $c_1,\dots,c_m$, then
\begin{align*}
{\coe}_{\node}({\bf x}) =  \sum_{i=1}^m \coe_{c_i}({\bf x}).
\end{align*}

\item[--] If $\node$ is an OR gate with (child node, wire parameter) inputs $(c_1,\theta_1),\dots, (c_m, \theta_m)$, then
\begin{align*}
{\coe}_{\node}({\bf x}) =    \sum_{i=1}^m  f(n, {\bf x}, c_i) \cdot \left({\coe}_{c_i}({\bf x}) + \theta_i\right).
\end{align*}
\end{itemize}
At root node $r$ with weight function $\coe_r({\bf x})$, the logistic circuit defines the posterior distribution on class variable $Y$~as
\begin{align}
\label{equation: probability}
{\Pr} ( Y = 1 \mid {\bf x}) = \frac{1}{1 + \exp\left(-\coe_{r}({\bf x})  \right)}.
\end{align}
\end{definition}
Using Boolean circuit flow, this definition essentially collects all the parameters on wires with flow 1 that reach the root, in order to then make a prediction.
This is illustrated in Figure~\ref{fig: logistic circuit} by coloring red the gates and wires whose parameters and weight function are propagated upward for the example assignment $A=0$, $B=1$, $C=1$, $D=0$.
The logistic circuit in Figure~\ref{fig: logistic circuit} defines the same posterior predictions as the table in Figure~\ref{fig: posterior distribution}. Specifically, for the example assignment, the weight function simply sums the parameters colored in red: $-5.8+2.3+3.9+1.5 = 1.9$. We then apply the logistic function (Eq.~\ref{equation: probability}) to get the classification probability $\Pr(Y=1 \mid \sample) = \frac{1}{1+\exp(-1.9)} = 86.99\%$.

\subsection{Real-Valued Data}
\label{s: real-valued data}
The semantics given so far assume Boolean inputs $\sample$, which is a rather restrictive assumption and prohibits many machine learning applications.
Next, we augment the logistic circuit semantics such that they can classify examples with continuous variables.

We interpret real-valued variables $q \in [0,1]$ as parameterizing an (independent) Bernoulli distribution (cf.~\citet{semanticLoss}). Each continuous variable represents the probability of the corresponding Boolean random variable $X$. For example, with $\bf q$ setting $A=0.4$, $B=0.8$, $C=0.2$, and $D=0.7$, the probability of $\neg A \land D$ would be $(1-0.4)\cdot 0.7=0.42$. The same distribution defines a probability for each logical sentence, and therefore each node in the logistic circuit.
This allows us to generalize Boolean flow as follows.

\begin{definition}[Probabilistic Circuit Flow]
\label{definition: probabilistic flow}
Consider a deterministic OR gate $n$.
Let $\bf q$ be a vector of probabilities, one for each variable in $\bf X$.
The probabilistic flow $f(n,{\bf q},c)$ of vector $\bf q$ between parent $n$ and child $c$ is 
\begin{align*}
    f(n,{\bf q},c) = {\Pr}_{\bf q}(c \mid n) = \frac{\Pr_{\bf q}(c \land n)}{ \Pr_{\bf q}(n) } = \frac{\Pr_{\bf q}(c)}{ \Pr_{\bf q}(n) },
\end{align*}
where $\Pr_{\bf q}(.)$ is the fully-factorized distribution where each variable in  $\bf X$ has the probability assigned by $\bf q$.
\end{definition}
Logistic circuit semantics now support continuous data (after normalizing to $[0,1]$), simply by replacing Boolean flow with probabilistic flow in Definition~\ref{de: circuit semantics}.
Note that probabilistic circuit flow has Boolean circuit flow as a special case, when ${\bf q}$ happens to be binary. 
Furthermore, due to the determinism and decomposability properties, the probabilities in Definition~\ref{definition: probabilistic flow} can be computed efficiently, together with all probabilistic circuit flows and weight functions in the logistic circuit. We defer the discussion of these computational details to Section~\ref{section: computing flows}.  
In the rest of this paper, we will abuse notation and have $\sample$ refer to Boolean inputs as well as continuous inputs~${\bf q}$ interchangeably.

\section{Parameter Learning}
\label{section: parameter learning}

A natural next question is how to learn logistic circuit parameters from complete data, for a fixed given circuit structure (structure learning is discussed in Section~\ref{s:structurelearning}). Furthermore, we ask whether those learned parameters are guaranteed to be optimal, globally minimizing a loss function.
We address these questions by showing how parameter learning can be reduced to logistic regression on a modified set of features, owing to logistic circuits' strong properties.

\subsection{Special Cases}

Before presenting the general reduction, we briefly discuss two special cases that establish some intuition.

\subsubsection{Linear Weight Functions} 
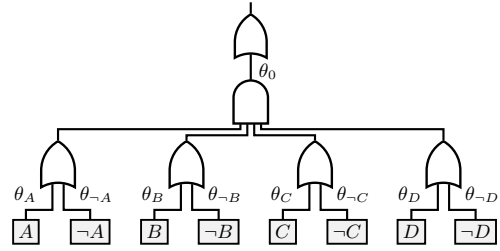
\begin{figure}[t]
\centering
 \begin{minipage}{0.48\textwidth}
 \centering
\scalebox{0.90}{
\begin{tikzpicture}[circuit logic US, nnf]
        
  \def\lvl{30pt}
  \def\ddhspace{15pt}
    
  \node (output) [] at (0,0){};%{Output};
  
  \node (root) [nnf3or] at ($(output) + (0pt,-0.7*\lvl)$){};%{r};
  
  \node (a1) [nnf4and] at ($(root) + (0pt,-1*\lvl)$){};%{a1};
  
  \node (o1a) [nnf2or] at ($(a1) + (-6*\ddhspace,-1*\lvl)$){};
  \node (o1b) [nnf2or] at ($(a1) + (-2*\ddhspace,-1*\lvl)$){};
  \node (o1c) [nnf2or] at ($(a1) + (2*\ddhspace,-1*\lvl)$){};
  \node (o1d) [nnf2or] at ($(a1) + (6*\ddhspace,-1*\lvl)$){};

  \node (ta1) [nnfterm] at ($(o1a) + (-1*\ddhspace,-1*\lvl)$){$A$};
  \node (ta0) [nnfterm] at ($(o1a) + (1*\ddhspace,-1*\lvl)$){$\neg A$};
  
  \node (tb1) [nnfterm] at ($(o1b) + (-1*\ddhspace,-1*\lvl)$){$B$};
  \node (tb0) [nnfterm] at ($(o1b) + (1*\ddhspace,-1*\lvl)$){$\neg B$};
  
  \node (tc1) [nnfterm] at ($(o1c) + (-1*\ddhspace,-1*\lvl)$){$C$};
  \node (tc0) [nnfterm] at ($(o1c) + (1*\ddhspace,-1*\lvl)$){$\neg C$};
  
  \node (td1) [nnfterm] at ($(o1d) + (-1*\ddhspace,-1*\lvl)$){$D$};
  \node (td0) [nnfterm] at ($(o1d) + (1*\ddhspace,-1*\lvl)$){$\neg D$};
  
  \begin{scope}[on background layer]
    \draw [nnfedge] (output) -- (root.output);
    \draw [nnfedge] (a1.output) -- (root.input 2) 
                        node[pos=0.32,right]  {$\theta_0$};
                        
    \draw [nnfedge] (o1a.output) -- ++(up:0.18) -| (a1.input 1);
    \draw [nnfedge] (o1b.output) -- ++(up:0.08) -| (a1.input 2);
    \draw [nnfedge] (o1c.output) -- ++(up:0.08) -| (a1.input 3);
    \draw [nnfedge] (o1d.output) -- ++(up:0.18) -| (a1.input 4);
    
    \draw [nnfedge] (ta1.north) -- ++(up:0.15) -| (o1a.input 1) 
                        node[pos=0.35,above left]  {$\theta_A$};
    \draw [nnfedge] (ta0.north) -- ++(up:0.15) -| (o1a.input 2) 
                        node[pos=0.35,above right]  {$\theta_{\neg A}$};
                        
    \draw [nnfedge] (tb1.north) -- ++(up:0.15) -| (o1b.input 1) 
                        node[pos=0.35,above left]  {$\theta_B$};
    \draw [nnfedge] (tb0.north) -- ++(up:0.15) -| (o1b.input 2) 
                        node[pos=0.35,above right]  {$\theta_{\neg B}$};
                        
    \draw [nnfedge] (tc1.north) -- ++(up:0.15) -| (o1c.input 1) 
                        node[pos=0.35,above left]  {$\theta_C$};
    \draw [nnfedge] (tc0.north) -- ++(up:0.15) -| (o1c.input 2) 
                        node[pos=0.35,above right]  {$\theta_{\neg C}$};
                        
    \draw [nnfedge] (td1.north) -- ++(up:0.15) -| (o1d.input 1) 
                        node[pos=0.35,above left]  {$\theta_D$};
    \draw [nnfedge] (td0.north) -- ++(up:0.15) -| (o1d.input 2) 
                        node[pos=0.35,above right]  {$\theta_{\neg D}$};
    
  \end{scope}
      
\end{tikzpicture}
}
\caption{Logistic regression represented as a logistic circuit}
\label{circuit: linear weights}
\end{minipage}
\end{figure}

Consider a vanilla logistic regression model on input variables (features) $\bf X$. 
Does there exist an equivalent logistic circuit with the same weight function?
For sample $\sample$, logistic regression with parameters ${\bm \theta}$ would have weight function $\sample \cdot {\bm \theta}$. 
Following Definition~\ref{de: circuit semantics}, we obtain such a simple weight function (linear in the input variables) by placing OR gates over complementary pairs of literals and associating a $\theta$ parameter which each wire (see Figure~\ref{circuit: linear weights}).\footnote{The negated variable inputs and parameters $\theta_{\neg X}$ are redundant, but we keep them for the sake of consistency. Alternatively, we can fix $\theta_{\neg X} = 0$ for all $X$ to remove this redundancy.} A large parent AND gate collects these variable-wise weights into a single linear sum. Finally, an OR gate at the root adds the bias term regardless of the~input.
\begin{proposition}
For each classical logistic regression model, there exists an equivalent logistic circuit model.
\end{proposition}

\subsubsection{Boolean Flow Indicators}
Next, let us consider a special case that makes no assumptions about circuit structure, but that requires the inputs to be fully binary. 
Such a circuit would have Boolean flows through every wire. 
Instead of working with the input variables $\bf X$, we can introduce new features that are indicator variables, telling us how the example propagates through the circuit, and which wires have a Boolean flow that reaches the circuit root. The circuit flows (indicators) decide which parameters are summed into the weight function; this process has been implicitly revealed in Figure~\ref{fig: logistic circuit}. By introducing such indicators, we can always obtain a linear weight function of composite features that are extracted from sample $\sample$. 
Next, we generalize this idea of introducing wire features to arbitrary logistic circuits.

\subsection{Reduction to Logistic Regression}
We will now consider the most general case, with continuous input data and no assumptions on the circuit structure.

\begin{proposition} 
\label{proposition: logistic regression}
Any logistic circuit model can be reduced to a logistic regression model over a particular feature~set.
\end{proposition}

\begin{corollary}
Logistic circuit cross-entropy loss is convex.
\end{corollary}

To prove Proposition~\ref{proposition: logistic regression}, we need to rewrite the classification distribution in Definition~\ref{de: circuit semantics} as follows.
$$
{\Pr} ( Y = 1 \mid {\bf x}) = \frac{1}{1+ \exp(- \bm{\mathbbm{x}} \cdot {\bm \theta})}.
$$ 
Here, $\bm{\mathbbm{x}}$ is some vector of features extracted from the raw example $\bf x$. This feature vector can only depend on $\bf x$; not on the parameters $\bm \theta$. 
Thus, the fundamental question is whether we can decompose $\coe_n(\sample)$ into $\bm{\mathbbm{x}} \cdot {\bf \theta}$ for all nodes $n$. We prove this to be true by induction:
\begin{itemize}
\item[--] \underline{Base case}: $\node$ is a leaf (input) node. It is obvious $\coe_n$ can be expressed as $\bm{\mathbbm{x}} \cdot {\bf \theta}$ since $\coe_n$ always equals 0.

\item[--] \underline{Induction step}: assume $\coe$ of all the nodes under node $n$ can be expressed as $\bm{\mathbbm{x}} \cdot {\bf \theta}$. We need to consider two cases:
\begin{enumerate}[wide=0pt, leftmargin=\dimexpr\labelwidth + 2\labelsep\relax]

\item If $\node$ is an AND gate having (w.l.o.g.) two children, prime $p$ and sub $s$. Given $\coe_{p} = \bm{\mathbbm{x}}_{p} \cdot \theta_{p}$ and $\coe_{s} =\bm{\mathbbm{x}}_s \cdot \theta_s$, 
\begin{align*}
\coe_\node &=  \bm{\mathbbm{x}}_{p} \cdot \theta_{p} + \bm{\mathbbm{x}}_s \cdot \theta_s \\
%&= [\bm{\mathbbm{x}}_p \bm{\mathbbm{x}}_s] \cdot [\theta_{p} \theta_s]
&= \begin{bmatrix} 
\bm{\mathbbm{x}}_p \\
\bm{\mathbbm{x}}_s
\end{bmatrix}
\cdot
\begin{bmatrix}
\theta_p \\
\theta_s
\end{bmatrix}.
\end{align*}

\item If $\node$ is an OR gate with (child node, wire parameter) inputs $\left\{(c_1,\theta_1),\dots,(c_m,\theta_m)\right\}$. Given $\coe_{c_i} =  \bm{\mathbbm{x}}_{c_i} \cdot \theta_{c_i}$,
\begin{align*} 
\coe_\node & = \sum_i  f(n,{\bf x},c_i) \cdot \left( \bm{\mathbbm{x}}_{c_i} \cdot \theta_{c_i}+ \theta_i\right) \\
&= \begin{bmatrix}
 f(n,{\bf x},c_1) \cdot\bm{\mathbbm{x}}_{c_1} \\
 f(n,{\bf x},c_1) \\
\vdots \\
 f(n,{\bf x},c_m) \cdot\bm{\mathbbm{x}}_{c_m} \\
 f(n,{\bf x},c_m) 
\end{bmatrix}
\cdot
 \begin{bmatrix}
\theta_{c_1} \\
\theta_1\\
\vdots \\
\theta_{c_m} \\
\theta_m
\end{bmatrix}.
\end{align*}
\end{enumerate}
\end{itemize}

Note that this proof holds true regardless of whether the input sample ${\bf x}$ is binary or real-valued.
With this proof, it is obvious that learning the parameters of a logistic circuit is equivalent to logistic regression on features $\mathbbm{x}$. 
We refer readers to \citet{Rennie2005} for a detailed proof that logistic regression is convex. 

Given this correspondence, any convex optimization technique can now be brought to bear on the problem of learning the parameters of a logistic circuit. In particular, we use stochastic gradient descent for this task.

\subsection{Global Circuit Flow Features}

In the proof of Proposition~\ref{proposition: logistic regression}, features $\mathbbm{x}$  are computed recursively by induction. However, it is not clear what these features represent, and how they are connected to the input data. In this section we assign semantics to those extracted features. They are the \emph{global circuit flow} of the observed example through the circuit. Global circuit flow is defined with respect to the root of a logistic circuit.

\begin{definition}[Global Circuit Flow]
    \label{definition: global flow}
    Consider a logistic circuit over variables $\bf X$ rooted at OR gate $r$.  
    The global circuit flow $f_r(n, \sample, c)$ of input $\sample $ between parent $n$ and child $c$ is defined inductively as follows.
    The global circuit flow between root $r$ and its child $c$ is the (local) probabilistic circuit flow: $f_r(r, \sample, c) = f(r, \sample, c)$.
    Then, for any node $n$ with parents $v_1,\dots,v_m$, we have that
    \begin{itemize}
        \item[--] if $n$ is an AND gate, global flow from child $c$ is
        \begin{align*}
            f_r(n, \sample, c) = \sum_{i=1}^m f_r(v_i, \sample, n),
        \end{align*}
        \item[--] if $n$ is an OR gate, global flow from child $c$ is
        \begin{align*}
            f_r(n, \sample, c) &=  f(n, \sample,c) \cdot \sum_{i=1}^m  f_r(v_i, \sample, n).
        \end{align*}
    \end{itemize}
\end{definition}
The red wires in Figure~\ref{fig: logistic circuit} have a global circuit flow of 1 for the given Boolean input. In general, global circuit flow assigns a continuous probability value to each wire.

Based on global circuit flow, we postulate the following alternative semantics for logistic circuits.
\begin{definition}[Logistic Circuit Alternative Semantics]
\label{definition: circuit semantic using global flow}
Let $\mathcal{W}$ be the set of all wires $(n,
\theta,c)$ between OR gates $n$ and children $c$ with parameters $\theta$.
Then, a logistic circuit rooted at node $r$ defines the weight function
$$
	\coe_r(\sample) = \sum_{(n,\theta,c) \in \mathcal{W}} f_r(n,\sample,c) \cdot \theta.
$$
\end{definition}
 
Note that the definition of global circuit flows, as well as our alternative semantics, follow a top-down induction. 
In contrasts, the original semantics in Definition~\ref{de: circuit semantics} follow a bottom-up induction.
We resolve this discrepancy next.
\begin{proposition}
\label{proposition: features}
The features $\mathbbm{x}$ constructed in the proof of Proposition~\ref{proposition: logistic regression} are equivalent to global flows $f_r(n,\sample,c)$.
\end{proposition}
\begin{corollary}
The bottom-up semantics of Definition~\ref{de: circuit semantics} and the top-down semantics of Definition~\ref{definition: circuit semantic using global flow} are equivalent.
\end{corollary}
\noindent We defer the proof of this proposition to Appendix~\ref{section: proof of proposition}.

Recall that without parameters, a logistic circuit is simply a logical circuit, which means that gates in a logistic circuit have real meaning: they correspond to some logical sentence. Hence, the values of global circuit flow features~$\mathbbm{x}$ correspond to probabilities of these logical sentences according to the input vector $\sample$. This provides us with a precious opportunity to assign meaning to the features learned by logistic circuits. We will revisit this point in Section~\ref{s: interpretability}, where we also visualize some global circuit flow features.

\subsection{Computing Global Flow Features Efficiently}
\label{section: computing flows}
While logistic circuit parameter learning is convex, we would like to also guarantee that the required feature computation is tractable. This section discusses efficient methods to calculate global flow features $\bm{\mathbbm{x}}$ (i.e., $f_r(n, \sample, c)$) from training samples $\bf x$ offline, before parameter learning. 
 
As is clear from Definition~\ref{definition: probabilistic flow}, circuit flows make extensive use of node probabilities.
We design our computation to consist of two parts, and dedicate the first part to the calculation of node probabilities. The first part is a bottom-up linear pass over the circuit starting with leaf nodes whose probabilities are directly provided by the input sample; see the details in Appendix~\ref{section: node probabilities}.
The second part makes use of these node probabilities to calculate the global circuit flow features in linear time. 
It is a top-down implementation of the recursion in Definition~\ref{definition: global flow}; see its details in Appendix~\ref{s: calculation of global flows}. 
Note that these computations correspond to the partial derivative computations used in arithmetic circuits for the purpose of probabilistic inference~\cite{DarwicheJACM}.

Our algorithm is completely compatible with fast vector arithmetic: instead of inputting one single sample each time,  one can directly supply the algorithms with a vector of samples (e.g., a mini batch), and this yields significant speedups.

\section{Structure Learning} \label{s:structurelearning}
This section presents an algorithm to learn a compact logical circuit structure for logistic circuits from data. 
For simplicity of designing the primitive operations, we assume AND gates always have two inputs (prime and sub).

\subsection{Learning Primitive}
The split operation was first introduced to modify the structure of PSDD circuits \cite{Liang2017}. We adopt it here with minor changes\footnote{Compared to the splits in LearnPSDD~\cite{Liang2017}, we do not limit constraints to be on primes.} as the primitive operation for our structure learning algorithm. 
Splitting an AND gate happens by imposing two additional constraints that are \emph{mutually exclusive} and \emph{exhaustive}, in particular by making two opposing variable assignments.
Executing a split creates partial copies of the gate and some of its decedents. Furthermore, one can choose to duplicate additional nodes up to a fixed depth (3 in our experiments).
We refer readers to \citet{Liang2017} for further details on the algorithm for executing splits.

\begin{figure}[t]	
\centering
	\begin{subfigure}[t]{0.22\textwidth}
	\centering	
    	\scalebox{0.85}{
	\begin{tikzpicture}[circuit logic US, nnf]
        
  \def\lvl{30pt}
      
  \node (output) [] at (455pt,1.88*\lvl){};
  
  \node (root) [nnf2or] at ($($(output) + (0pt,-0.7*\lvl)$)$){};
  
  \node (a1) [nnf2and] at ($(root) + (-20pt,-1.0*\lvl)$){};
  \node (a2) [nnf2and] at ($(root) + (20pt,-1.0*\lvl)$){};
  
  \node (o12) [nnf2or] at ($(root) + (0pt,-1.95*\lvl)$){};
  
  \node (ta1) [nnfterm] at ($(o12) + (-10pt,-0.7*\lvl)$){$A$};
  \node (ta0) [nnfterm] at ($(o12) + (10pt,-0.7*\lvl)$){$\neg A$};
  
  \node (tb1) [nnfterm] at ($(a1) + (-2.8pt,-0.8*\lvl)$){$B$};
  \node (tb0) [nnfterm] at ($(a2) + (2.8pt,-0.8*\lvl)$){$\neg B$};
  
  \begin{scope}[on background layer]
  
    \draw [nnfedge] (output) -- (root.output);
    
    \draw [nnfedge] (ta1.north) -- ++(up:0.10) -| (o12.input 1);
    \draw [nnfedge] (ta0.north) -- ++(up:0.10) -| (o12.input 2);
    
    \draw [nnfedge] (tb1.north) -- (a1.input 1);
    \draw [nnfedge] (tb0.north) -- (a2.input 2);
    
    \draw [nnfedge] (o12.output) -- ++(up:0.10) -| (a1.input 2);
    \draw [nnfedge] (o12.output) -- ++(up:0.10) -| (a2.input 1);
    
    \draw [nnfedge,red] (a1.output) -- ++(up:0.10) -| (root.input 1)
                        node[pos=0.3,above left,red]  {$f_0$};
    \draw [nnfedge] (a2.output) -- ++(up:0.10) -| (root.input 2);

  \end{scope}
\end{tikzpicture}
}
    	\caption{Before split of $f_0$ on $A$} \label{fig: split:before}
	\end{subfigure}
	~~~
	\begin{subfigure}[t]{0.22\textwidth}
	\centering	
    	\scalebox{0.85}{
	\begin{tikzpicture}[circuit logic US, nnf]
        
  \def\lvl{30pt}
      
  \node (output) [] at (455pt,1.88*\lvl){};
  
  \node (root) [nnf3or] at ($($(output) + (0pt,-0.7*\lvl)$)$){};
  
  \node (a11) [nnf2and] at ($(root) + (-50pt,-1.0*\lvl)$){};%{a1};
  \node (a12) [nnf2and] at ($(root) + (-20pt,-1.0*\lvl)$){};%{a1};
  \node (a2) [nnf2and] at ($(root) + (20pt,-1.0*\lvl)$){};%{a2};
  
  \node (o12) [nnf2or] at ($(root) + (0pt,-1.95*\lvl)$){};%{o12};
  
  \node (ta1) [nnfterm] at ($(o12) + (-10pt,-0.7*\lvl)$){$A$};
  \node (ta0) [nnfterm] at ($(o12) + (10pt,-0.7*\lvl)$){$\neg A$};
  \node (ta0c) [nnfterm] at ($(o12) + (-52.8pt,-0.7*\lvl)$){$\neg A$};
  
  \node (tb1) [nnfterm] at ($(a1) + (-9pt,-1.65*\lvl)$){$B$};
  \node (tb0) [nnfterm] at ($(a2) + (2.8pt,-0.8*\lvl)$){$\neg B$};
  
  \begin{scope}[on background layer]
  
    \draw [nnfedge] (output) -- (root.output);
    
    \draw [nnfedge] (ta1.north) -- ++(up:0.10) -| (o12.input 1);
    \draw [nnfedge] (ta0.north) -- ++(up:0.10) -| (o12.input 2);
    
    \draw [nnfedge] (tb0.north) -- (a2.input 2);
    
    \draw [nnfedge] (tb1.north) -- ++(up:0.50) -| (a12.input 1);
    \draw [nnfedge] (tb1.north) -- ++(up:0.50) -| (a11.input 2);
    
    \draw [nnfedge] (ta0c.north) -- ++(up:0.10) -| (a11.input 1);
    \draw [nnfedge] (ta1.north) -- ++(up:0.10) -| (a12.input 2);
    
    \draw [nnfedge] (o12.output) -- ++(up:0.10) -| (a2.input 1);
    
    \draw [nnfedge,red] (a11.output) -- ++(up:0.20) -| (root.input 1)
                        node[pos=0.3,above left,red]  {$f_1$};
    \draw [nnfedge,red] (a12.output) -- ++(up:0.10) -| (root.input 2)
                        node[pos=0.2,below right,red]  {$f_2$};
    \draw [nnfedge] (a2.output) -- ++(up:0.10) -| (root.input 3);

  \end{scope}
\end{tikzpicture}
	}
    	\caption{After split of $f_0$ on $A$} \label{fig: split:after}
	\end{subfigure}
	\\[5pt]
    \begin{subfigure}[t]{0.48\textwidth}
    \centering
    \begin{sc}
    {\fontsize{9}{9}\selectfont
            \begin{tabular}{ @{} ll  c  c c cc@{} }
             \toprule
              $A$ & $B$  & & $f_0$ & & $f_1$ & $f_2$ \\
             \midrule \midrule
             1 & 1 & &1 & & 0 & 1 \\
             0 & 1 & &1 & & 1 & 0 \\
             \midrule
    	0.5 & 0.6 & & 0.6 & & 0.30 & 0.30 \\
    	0.4 & 0.8 & & 0.8 & & 0.48 & 0.32 \\
              \bottomrule
           	\end{tabular}
    	}
              \end{sc}
    		\caption{Circuit flow before and after the split.}\label{table: flow change}
	\end{subfigure}
	\caption{A split changes the circuit flow.}\label{fig: split with flow change}
\end{figure}
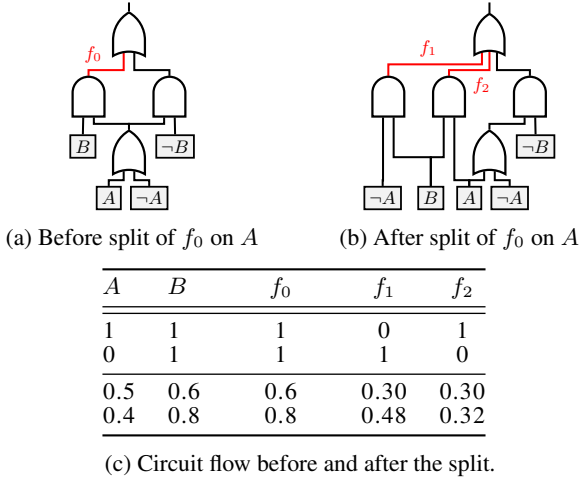

Splits are ideal primitives to change the classifier induced by a logistic circuit: they directly affect the circuit flows (see Figure~\ref{fig: split with flow change}). By imposing constraints on AND gates, splits alter the node probabilities associated with the affected AND gates. Following Definition~\ref{definition: probabilistic flow}, the circuit flows on the wires out of those AND gates adapt accordingly. While Figure~\ref{fig: split with flow change} focuses on the immediately affected wires, the effect of a split on circuit flows can propagate downward for several levels, depending on the depth of node duplication. Still the effects of a split on both the structure of a logistic circuit and the circuit flows are very local and contained in the sub-circuit rooted at the OR parent of the split AND gate. However, its effect on the parameters is global. Once a split is executed, the whole parameter set needs to be re-trained.

\subsection{Learning Algorithm}
The overall structure learning algorithm for logistic circuits, built on top of the split operation, proceeds as follows. Iteratively, one split is executed to change the structure, followed by parameter learning. 
We only consider single-variable split constraints and first select which AND gate to split, followed by a selection of which variable to split on.

When using gradient descent, one hopes that the parameter on the AND gate output consistently has its partial derivatives pointing in the same direction for all training examples. This will steadily push the parameter to a large magnitude. 

If this is not the case, we will use splits to alter the flow of examples through the circuit.
Specifically, those AND gates whose associated output parameter has a large variance of its partial derivative (that is, the derivative of the loss function w.r.t.~that parameter) requires splitting for the parameters to improve.  We simply select the AND gate whose output parameter has the highest training variance.

Given an AND gate to split, we consider candidate variables $X$ to execute the split with. We construct two sets of training examples that affect this node: in one group, each example is weighted by the marginal probability of $X$; in the other, 
with the marginal probability of $\neg X$.  
Next, we calculate the within-group weighted variances of the partial derivatives.
The variable with the smallest weighted variances gets picked, as this suggests the split will introduce new parameters with gradients that align in one direction.
   
\begin{table}[t]
 \centering
 \begin{minipage}{0.48\textwidth}
          \caption{Classification accuracy of logistic circuits in context with commonly used existing models. We report the details of those existing models in Appendix~\ref{s: model details}.} 
          \label{table: accuracy}
          \centering
          {\fontsize{8.5}{9}\selectfont
          \begin{sc}
          \begin{tabular}{ @{}l c c @{} }
          \toprule
       	Accuracy $\%$ on Dataset & Mnist & Fashion \\
          \midrule\midrule
           Baseline: Logistic Regression & 85.3 & 79.3 \\
           Baseline: Kernel Logistic Regression & 97.7 & 88.3 \\
           Random Forest & 97.3 & 81.6 \\
            3-layer MLP\label{3MLP}
 & 97.5 & 84.8\\
 	RAT-SPN \cite{rat-spn2018} &98.1 & 89.5\\
	SVM with RBF Kernel & 98.5 & 87.8 \\
	            5-Layer MLP  & 99.3 & 89.8 \\
            \midrule 
          Logistic Circuit (binary) &  97.4 & 87.6 \\
          Logistic Circuit (real-valued) & 99.4 & 91.3\\ 
          \midrule
      % CNN with 2 conv layers & 98.8  & 89.9 \\         
            CNN with 3 conv layers & 99.1  &90.7\\
            Resnet \cite{he2016cvpr}& 99.5 & 93.6 \\
		\bottomrule
		\end{tabular}
          \end{sc}
     }
     \end{minipage}     
   \end{table}
   
          \begin{table}[tb]
 \centering
  \begin{minipage}{0.48\textwidth}
     {\footnotesize
          \caption{Number of parameters of logistic circuits in context with existing SGD-based models, when achieving the classification accuracy reported in Table~\ref{table: accuracy}
          }
          \label{table: size}
          \centering
          {\fontsize{8.3}{9}\selectfont
          \begin{sc}
          \begin{tabular}{ @{}l r r @{} }
          \toprule
       	 Number of Parameters & Mnist & Fashion \\
          \midrule\midrule
           Baseline: Logistic Regression & $<$1K & $<$1K \\
           Baseline: Kernel Logistic Regression & 1,521 K & 3,930K\\
                       \midrule 
          Logistic Circuit (real-valued) & 182K & 467K\\
          Logistic Circuit (binary) & 268K & 614K \\
          \midrule
            3-layer MLP  & 1,411K  & 1,411K \\
 	RAT-SPN~ \cite{rat-spn2018} 	 & 8,500K & 650K \\
	CNN with 3 conv layers  & 2,196K & 2,196K\\
	            5-Layer MLP & 2,411K &  2,411K \\
                        Resnet \cite{he2016cvpr} & 4,838K & 4,838K \\
		\bottomrule
		\end{tabular}
          \end{sc}
     }}
    \end{minipage}
   \end{table}
     
   \begin{table*}[tb]
          \caption{Comparison of logistic circuits with MLPs when trained with different percentages of the dataset.}
          \label{table: data efficiency}
          \centering
          {\fontsize{9}{9}\selectfont
          \begin{sc}
          \begin{tabular}{ @{}l c c c c c c@{} }
          \toprule
        \multirow{2}{*}{Accuracy $\%$  with $\%$ of Training Data }& \multicolumn{3}{c}{MNIST} & \multicolumn{3}{c}{Fashion}\\
	\cmidrule{2-4} \cmidrule{5-7} 
	&100$\%$ & 10$\%$ & 2$\%$ & 100$\%$ & 10$\%$ & 2$\%$ \\
	          \midrule\midrule
	5-layer MLP & 99.3 & {\bf 98.2} &  94.3   & 89.8 & 86.5 & 80.9 \\ 
	CNN with 3 Conv Layers & 99.1 & 98.1 & 95.3 &90.7 & 87.6 & 83.8  \\
	\midrule
                   	Logistic Circuit (Binary) &  97.4 & 96.9 &  94.1 & 87.6 & 86.7 & 83.2 \\
         	Logistic Circuit (Real-Valued) &  {\bf 99.4} & 97.8 &  {\bf 96.1} & {\bf 91.3} & {\bf 87.8} & {\bf 86.0} \\
       		\bottomrule
		\end{tabular}
          \end{sc}
          }
   \end{table*}

\section{Empirical Evaluation} \label{s:experiments}

In this section, we empirically evaluate the competitiveness of our learner on three aspects: classification accuracy, model complexity,  and data efficiency.\footnote{Open-source code and experiments are available at 
\url{https://github.com/UCLA-StarAI/LogisticCircuit}.}
 Moreover, we visualize the most important active feature with regards to the given sample to provide local interpretation for why the learned logistic circuit makes such classification.
 \subsection{Setup \& Data Preprocessing}
 We choose MNIST and Fashion\footnote{A dataset of Zalando's images,  intended as a more challenging drop-in replacement of MNIST \cite{fashion2017}.} as our testbeds.
 Since logistic circuits are intended for binary classification, we use the standard ``one vs.~rest" approach to construct an ensemble multi-class classifier such that our method can be evaluated on these two datasets. When running the binary logistic circuit, we transform pixels that are smaller than their mean plus $0.05$ standard deviation to 0 and the rest to 1. When running the real-valued version, we transform pixels to $[0,1]$ by dividing them by 255. All experiments start with a predefined initial structure; we defer its details to Appendix~\ref{appendix: initial structure}.
  The learned structure with the highest F1 score on validation after 48 hours of running is used for evaluation. All experiments are run on single CPUs.

\begin{figure}[t]
    \centering
	\begin{subfigure}[t]{0.22\textwidth}
    	\centering	
	\includegraphics[width=0.95\textwidth]{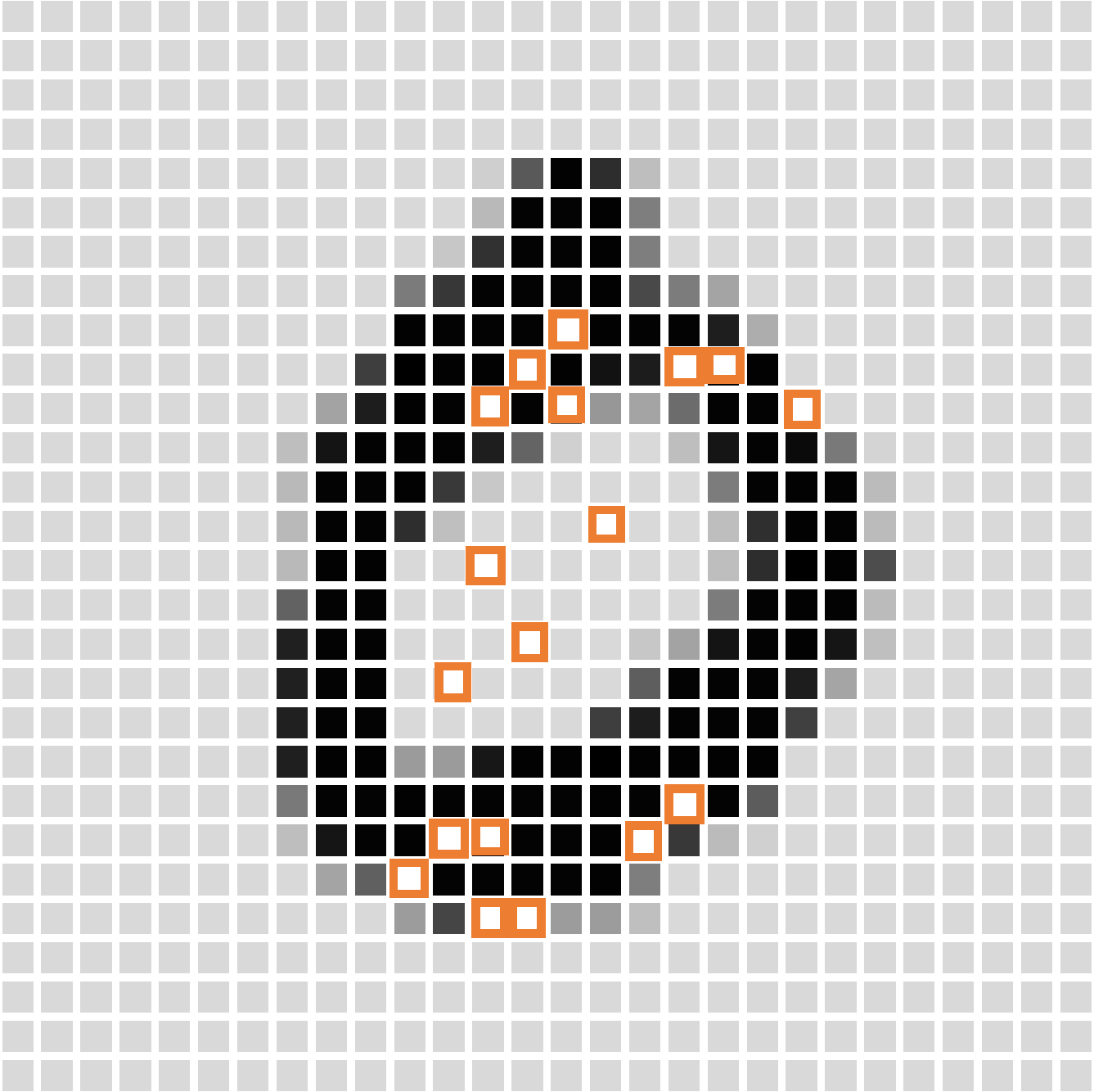} 
	\end{subfigure}
	~
	\begin{subfigure}[t]{0.22\textwidth}
	    \centering	
    	\includegraphics[width=0.95\textwidth]{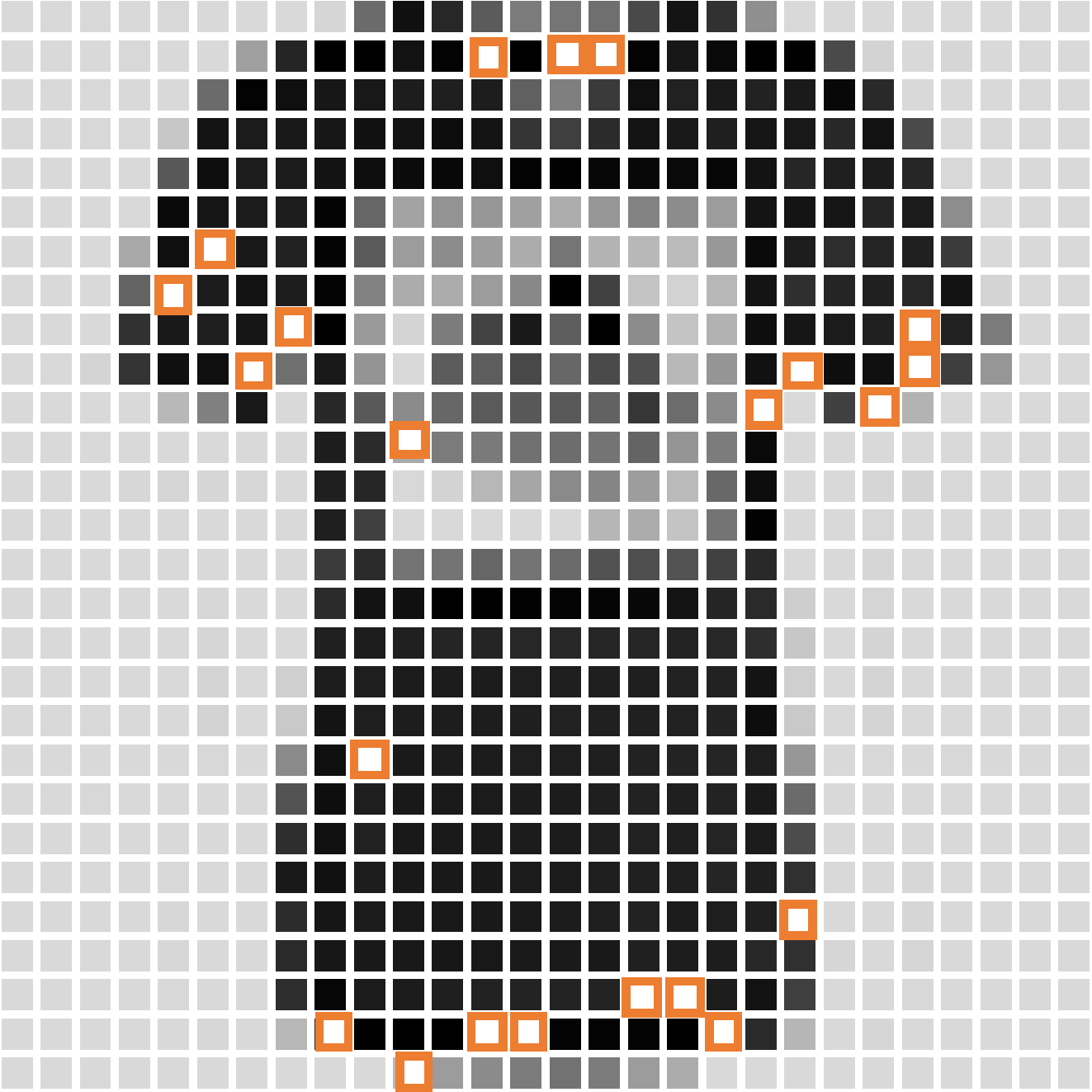}
		\end{subfigure}
\caption{Visualization of the single compositional feature that contributes most to the classification probability with regards to the input image. Features are marked in orange. Left: a digit 0 from MNIST. Right: a t-shirt from Fashion. }
\label{fig: feature visualization}
\end{figure}
   
\subsection{Classification Accuracy}
\label{s: classification accuracy}
Table~\ref{table: accuracy} summarizes the classification accuracy on test data. Learning a logistic circuit on the binary data is on par with a 3-layer MLP; the real-valued version outperforms 5-layer MLPs and even CNNs with 3 convolutional layers. The fact that logistic circuits achieve better accuracy than CNNs is surprising, since logistic circuits do not use convolutions, which are specifically designed to exploit image invariances. 

In addition, we would like to emphasis our comparison with two of the baselines. As parameter learning of logistic circuits is equivalent to logistic regression, one can view structure learning of logistic circuits as a process of constructing composite features from raw samples.  
The significant improvement over standard logistic regression demonstrates the effectiveness of our method in extracting valuable features; using kernel logistic regression can only partially bridge the gap in performance, yet as shown later, it does so at the cost of introducing many more parameters. 

We also want to call attention to our comparison with RAT-SPN, the current state of the art in discriminative learning for probabilistic circuits. SPN is another form of circuit representation, with less restrictive structure. Parameter learning in SPN is not convex and generally requires other techniques such as EM or non-convex optimization. The empirical observation that our method achieves significantly better classification accuracy than RAT-SPN demonstrates that in structure learning, imposing more restrictions on the model's structural syntax may be beneficial. The syntactic restriction of logistic circuits requires decomposability and determinism; without them, convex parameter learning does not appear to be possible. 
As structure learning is built on top of parameter learning, a well-behaved parameter learning loss with a unique optimum can provide more informative guidance about how to adapt the structure, leading to a more competitive structure learning algorithm overall.

\begin{figure*}[t]
    \centering
	\begin{subfigure}[t]{0.62\textwidth}
    	\centering	
    	\scalebox{0.7}{
	\begin{tikzpicture}[circuit logic US, nnf]
        
  \def\lvl{30pt}

  \node (output) [] at (105pt,1.88*\lvl){};%{Output};
  
  \node (root) [nnf2or] at ($($(output) + (0pt,-0.7*\lvl)$)$){};
  
  \node (ar1) [nnf2and] at ($(root) + (-60pt,-0.9*\lvl)$){};
  \node (ar2) [nnf2and] at ($(root) + (60pt,-0.9*\lvl)$){};
  
  \node (ty1) [nnfterm] at ($(ar1) + (20pt,-0.8*\lvl)$){$Y$};
  \node (ty0) [nnfterm] at ($(ar2) + (-20pt,-0.8*\lvl)$){$\neg Y$};
  
  \node (rootL) [nnf2or] at ($(0,0) + (0pt,-0.7*\lvl)$){};%{r};
  
  \node (a1) [nnf2and] at ($(rootL) + (-60pt,-1*\lvl)$){};
  \node (a2) [nnf2and] at ($(rootL) + (60pt,-1*\lvl)$){};
  
  \node (o12) [nnf2or] at ($(a1) + (20pt,-1*\lvl)$){};
  \node (o21) [nnf2or] at ($(a2) + (-20pt,-1*\lvl)$){};

  \node (aeq1) [nnf2and] at ($(rootL) + (-60pt,-3*\lvl)$){};
  \node (aneq) [nnf2and] at ($(rootL) + (0pt,-3*\lvl)$){};
  \node (aeq2) [nnf2and] at ($(rootL) + (60pt,-3*\lvl)$){};

  \node (oeq) [nnf2or] at ($(rootL) + (20pt,-4.6*\lvl)$){};
  \node (oneq) [nnf2or] at ($(rootL) + (80pt,-4.6*\lvl)$){};
  
  \node (se1) [nnf2or] at ($(rootL) + (-70pt,-4.7*\lvl)$){};
  \node (se2) [nnf2or] at ($(rootL) + (-30pt,-4.7*\lvl)$){};
  
  \node (aab11) [nnf2and] at ($(oeq) + (-14.8pt,-1*\lvl)$){};
  \node (aab00) [nnf2and] at ($(oeq) + (14.7pt,-1*\lvl)$){};
  \node (aab10) [nnf2and] at ($(oneq) + (-14.8pt,-1*\lvl)$){};
  \node (aab01) [nnf2and] at ($(oneq) + (15.2pt,-1*\lvl)$){};
  
  \node (tc1) [nnfterm] at ($(a1) + (-20pt,-.8*\lvl)$){$A$};
  \node (tc0) [nnfterm] at ($(a2) + (20pt,-.8*\lvl)$){$\neg A$};
  
  \node (te1) [nnfterm] at ($(rootL) + (-72.5pt,-6.1*\lvl)$){$B$};
  \node (te0) [nnfterm] at ($(rootL) + (-27.4pt,-6.1*\lvl)$){$\neg B$};
  
  \node (ta1) [nnfterm] at ($(rootL) + (2.5pt,-7*\lvl)$){$C$};
  \node (ta0) [nnfterm] at ($(rootL) + (32.1pt,-7*\lvl)$){$\neg C$};
  
  \node (tb1) [nnfterm] at ($(rootL) + (97.8pt,-7*\lvl)$){$D$};
  \node (tb0) [nnfterm] at ($(rootL) + (67.8pt,-7*\lvl)$){$\neg D$};
  
  \begin{scope}[on background layer]
  
    \draw [nnfedge] (output) -- (root.output);
    
    \draw [nnfedge] (ar1.output) -- ++(up:0.10) -| (root.input 1) 
                        node[pos=0.4,above left]  {$0.6$};
    \draw [nnfedge] (ar2.output) -- ++(up:0.10) -| (root.input 2) 
                        node[pos=0.4,above right]  {$0.4$};
  
    \draw [nnfedge] (rootL.output) -- ++(up:0.10) -| (ar1.input 1);
    \draw [nnfedge] (ty1.north) -- ++(up:0.15) -| (ar1.input 2);
    
    \draw [nnfedge] (a1.output) -- ++(up:0.15) -| (rootL.input 1) 
                        node[pos=0.4,above left]  {$0.9$};
    \draw [nnfedge] (a2.output) -- ++(up:0.15) -| (rootL.input 2)
                        node[pos=0.4,above right]  {$0.1$};
    \draw [nnfedge] (o12.output) -- ++(up:0.15) -| (a1.input 2);
    \draw [nnfedge] (o21.output) -- ++(up:0.15) -| (a2.input 1);
    \draw [nnfedge] (aeq1.output) -- ++(up:0.15) -| (o12.input 1)
                        node[pos=0.3,above left]  {$0.2$};
    \draw [nnfedge] (aneq.output) -- ++(up:0.15) -| (o12.input 2)
                        node[pos=0.4,above right]  {$0.8$};
    \draw [nnfedge] (aeq2.output) -- ++(up:0.15) -| (o21.input 2)
                        node[pos=0.3,above right]  {$0.4$};
    \draw [nnfedge] (aneq.output) -- ++(up:0.15) -| (o21.input 1)
                        node[pos=0.4,above left]  {$0.6$};
    \draw [nnfedge] (oeq.output) -- ++(up:0.15) -| (aeq1.input 2);
    \draw [nnfedge] (oeq.output) -- ++(up:0.15) -| (aeq2.input 2);
    \draw [nnfedge] (oneq.output) -- ++(up:0.65) -| (aneq.input 2);
    
    \draw [nnfedge] (se1.output) -- ++(up:0.27) -| (aeq1.input 1);
    \draw [nnfedge] (se2.output) -- ++(up:0.52) -| (aeq2.input 1);
    \draw [nnfedge] (se2.output) -- ++(up:0.52) -| (aneq.input 1);
    
    \draw [nnfedge] (aab11.output) -- ++(up:0.15) -| (oeq.input 1) 
                        node[pos=0.3,above left]  {$0.1$};
    \draw [nnfedge] (aab00.output) -- ++(up:0.15) -| (oeq.input 2) 
                        node[pos=0.3,above right]  {$0.9$};
    \draw [nnfedge] (aab10.output) -- ++(up:0.15) -| (oneq.input 1)
                        node[pos=0.3,above left]  {$0.3$};
    \draw [nnfedge] (aab01.output) -- ++(up:0.15) -| (oneq.input 2)
                        node[pos=0.3,above right]  {$0.7$};
    
    \draw [nnfedge] (tc1.north) -- ++(up:0.15) -| (a1.input 1);
    \draw [nnfedge] (tc0.north) -- ++(up:0.15) -| (a2.input 2);
    \draw [nnfedge] (te1.north) -- ++(up:0.15) -| (se1.input 1)
                        node[pos=0.65,above left]  {$0.1$};
    \draw [nnfedge] (te0.north) -- ++(up:0.40) -| (se1.input 2)
                        node[pos=0.52,above right]  {$0.9$};
    \draw [nnfedge] (te1.north) -- ++(up:0.15) -| (se2.input 1)
                        node[pos=0.66,above left]  {$0.8$};
    \draw [nnfedge] (te0.north) -- ++(up:0.40) -| (se2.input 2)
                        node[pos=0.47,above right]  {$0.2$};
    
    \draw [nnfedge] (ta1.north) -- ++(up:0.15) -| (aab11.input 1);
    \draw [nnfedge] (ta1.north) -- ++(up:0.15) -| (aab10.input 1);
    \draw [nnfedge] (ta0.north) -- ++(up:0.35) -| (aab01.input 1);
    \draw [nnfedge] (ta0.north) -- ++(up:0.35) -| (aab00.input 1);
    
    \draw [nnfedge] (tb1.north) -- ++(up:0.55) -| (aab11.input 2);
    \draw [nnfedge] (tb1.north) -- ++(up:0.55) -| (aab01.input 2);
    \draw [nnfedge] (tb0.north) -- ++(up:0.75) -| (aab00.input 2);
    \draw [nnfedge] (tb0.north) -- ++(up:0.75) -| (aab10.input 2);
    
  \end{scope}

  \node (rootR) [nnf2or] at ($(210pt,0) + (0pt,-0.7*\lvl)$){};%{r};
  
  \node (a1) [nnf2and] at ($(rootR) + (-60pt,-1*\lvl)$){};%{a1};
  \node (a2) [nnf2and] at ($(rootR) + (60pt,-1*\lvl)$){};%{a2};
  
  \node (o12) [nnf2or] at ($(a1) + (20pt,-1*\lvl)$){};%{o12};
  \node (o21) [nnf2or] at ($(a2) + (-20pt,-1*\lvl)$){};%{o21};

  \node (aeq1) [nnf2and] at ($(rootR) + (-60pt,-3*\lvl)$){};%{aeq1};
  \node (aneq) [nnf2and] at ($(rootR) + (0pt,-3*\lvl)$){};%{aneq};
  \node (aeq2) [nnf2and] at ($(rootR) + (60pt,-3*\lvl)$){};%{aeq2};

  \node (oeq) [nnf2or] at ($(rootR) + (20pt,-4.6*\lvl)$){};%{oeq};
  \node (oneq) [nnf2or] at ($(rootR) + (80pt,-4.6*\lvl)$){};%{oneq};
  
  \node (se1) [nnf2or] at ($(rootR) + (-70pt,-4.7*\lvl)$){};%{se1};
  \node (se2) [nnf2or] at ($(rootR) + (-30pt,-4.7*\lvl)$){};%{se2};
  
  \node (aab11) [nnf2and] at ($(oeq) + (-14.8pt,-1*\lvl)$){};%{aab11};
  \node (aab00) [nnf2and] at ($(oeq) + (14.7pt,-1*\lvl)$){};%{aab00};
  \node (aab10) [nnf2and] at ($(oneq) + (-14.8pt,-1*\lvl)$){};%{aab10};
  \node (aab01) [nnf2and] at ($(oneq) + (15.2pt,-1*\lvl)$){};%{aab01};
  
  \node (tc1) [nnfterm] at ($(a1) + (-20pt,-.8*\lvl)$){$A$};
  \node (tc0) [nnfterm] at ($(a2) + (20pt,-.8*\lvl)$){$\neg A$};
  
  \node (te1) [nnfterm] at ($(rootR) + (-72.5pt,-6.1*\lvl)$){$B$};
  \node (te0) [nnfterm] at ($(rootR) + (-27.4pt,-6.1*\lvl)$){$\neg B$};
  
  \node (ta1) [nnfterm] at ($(rootR) + (2.5pt,-7*\lvl)$){$C$};
  \node (ta0) [nnfterm] at ($(rootR) + (32.1pt,-7*\lvl)$){$\neg C$};
  
  \node (tb1) [nnfterm] at ($(rootR) + (97.8pt,-7*\lvl)$){$D$};
  \node (tb0) [nnfterm] at ($(rootR) + (67.8pt,-7*\lvl)$){$\neg D$};
  
  \begin{scope}[on background layer]
    
    \draw [nnfedge] (rootR.output) -- ++(up:0.10) -| (ar2.input 2);
    \draw [nnfedge] (ty0.north) -- ++(up:0.15) -| (ar2.input 1);
    
    \draw [nnfedge] (a1.output) -- ++(up:0.15) -| (rootR.input 1) 
                        node[pos=0.4,above left]  {$0.4$};
    \draw [nnfedge] (a2.output) -- ++(up:0.15) -| (rootR.input 2)
                        node[pos=0.4,above right]  {$0.6$};
    \draw [nnfedge] (o12.output) -- ++(up:0.15) -| (a1.input 2);
    \draw [nnfedge] (o21.output) -- ++(up:0.15) -| (a2.input 1);
    \draw [nnfedge] (aeq1.output) -- ++(up:0.15) -| (o12.input 1)
                        node[pos=0.3,above left]  {$0.2$};
    \draw [nnfedge] (aneq.output) -- ++(up:0.15) -| (o12.input 2)
                        node[pos=0.4,above right]  {$0.8$};
    \draw [nnfedge] (aeq2.output) -- ++(up:0.15) -| (o21.input 2)
                        node[pos=0.3,above right]  {$0.3$};
    \draw [nnfedge] (aneq.output) -- ++(up:0.15) -| (o21.input 1)
                        node[pos=0.4,above left]  {$0.7$};
    \draw [nnfedge] (oeq.output) -- ++(up:0.15) -| (aeq1.input 2);
    \draw [nnfedge] (oeq.output) -- ++(up:0.15) -| (aeq2.input 2);
    \draw [nnfedge] (oneq.output) -- ++(up:0.65) -| (aneq.input 2);
    
    \draw [nnfedge] (se1.output) -- ++(up:0.27) -| (aeq1.input 1);
    \draw [nnfedge] (se2.output) -- ++(up:0.52) -| (aeq2.input 1);
    \draw [nnfedge] (se2.output) -- ++(up:0.52) -| (aneq.input 1);
    
    \draw [nnfedge] (aab11.output) -- ++(up:0.15) -| (oeq.input 1) 
                        node[pos=0.3,above left]  {$0.8$};
    \draw [nnfedge] (aab00.output) -- ++(up:0.15) -| (oeq.input 2) 
                        node[pos=0.3,above right]  {$0.2$};
    \draw [nnfedge] (aab10.output) -- ++(up:0.15) -| (oneq.input 1)
                        node[pos=0.3,above left]  {$0.5$};
    \draw [nnfedge] (aab01.output) -- ++(up:0.15) -| (oneq.input 2)
                        node[pos=0.3,above right]  {$0.5$};
    
    \draw [nnfedge] (tc1.north) -- ++(up:0.15) -| (a1.input 1);
    \draw [nnfedge] (tc0.north) -- ++(up:0.15) -| (a2.input 2);
    \draw [nnfedge] (te1.north) -- ++(up:0.15) -| (se1.input 1)
                        node[pos=0.65,above left]  {$0.6$};
    \draw [nnfedge] (te0.north) -- ++(up:0.40) -| (se1.input 2)
                        node[pos=0.52,above right]  {$0.4$};
    \draw [nnfedge] (te1.north) -- ++(up:0.15) -| (se2.input 1)
                        node[pos=0.66,above left]  {$0.9$};
    \draw [nnfedge] (te0.north) -- ++(up:0.40) -| (se2.input 2)
                        node[pos=0.47,above right]  {$0.1$};
    
    \draw [nnfedge] (ta1.north) -- ++(up:0.15) -| (aab11.input 1);
    \draw [nnfedge] (ta1.north) -- ++(up:0.15) -| (aab10.input 1);
    \draw [nnfedge] (ta0.north) -- ++(up:0.35) -| (aab01.input 1);
    \draw [nnfedge] (ta0.north) -- ++(up:0.35) -| (aab00.input 1);
    
    \draw [nnfedge] (tb1.north) -- ++(up:0.55) -| (aab11.input 2);
    \draw [nnfedge] (tb1.north) -- ++(up:0.55) -| (aab01.input 2);
    \draw [nnfedge] (tb0.north) -- ++(up:0.75) -| (aab00.input 2);
    \draw [nnfedge] (tb0.north) -- ++(up:0.75) -| (aab10.input 2);
    
  \end{scope}
      
\end{tikzpicture}
	}
    	\caption{Probabilistic circuit for joint distribution $\Pr(Y,A,B,C,D)$} \label{circuit: prop circuit: prob }
	\end{subfigure}
	~~~~~
	\begin{subfigure}[t]{0.34\textwidth}
	    \centering	
    	\scalebox{0.7}{
	\begin{tikzpicture}[circuit logic US, nnf]
        
  \def\lvl{30pt}
      
  \node (output) [] at (455pt,1.88*\lvl){};%{Output};
  
  \node (root) [nnf3or] at ($($(output) + (0pt,-0.7*\lvl)$)$){};
  
  \node (rootLL) [nnf2or] at ($(0,0) + (455pt,-0.7*\lvl)$){};%{r};
  
  \node (a1) [nnf2and] at ($(rootLL) + (-60pt,-1*\lvl)$){};
  \node (a2) [nnf2and] at ($(rootLL) + (60pt,-1*\lvl)$){};
  
  \node (o12) [nnf2or] at ($(a1) + (20pt,-1*\lvl)$){};
  \node (o21) [nnf2or] at ($(a2) + (-20pt,-1*\lvl)$){};

  \node (aeq1) [nnf2and] at ($(rootLL) + (-60pt,-3*\lvl)$){};
  \node (aneq) [nnf2and] at ($(rootLL) + (0pt,-3*\lvl)$){};
  \node (aeq2) [nnf2and] at ($(rootLL) + (60pt,-3*\lvl)$){};

  \node (oeq) [nnf2or] at ($(rootLL) + (20pt,-4.6*\lvl)$){};
  \node (oneq) [nnf2or] at ($(rootLL) + (90pt,-4.6*\lvl)$){};
  
  \node (se1) [nnf2or] at ($(rootLL) + (-80pt,-4.7*\lvl)$){};
  \node (se2) [nnf2or] at ($(rootLL) + (-35pt,-4.7*\lvl)$){};
  
  \node (aab11) [nnf2and] at ($(oeq) + (-14.8pt,-1*\lvl)$){};
  \node (aab00) [nnf2and] at ($(oeq) + (14.7pt,-1*\lvl)$){};
  \node (aab10) [nnf2and] at ($(oneq) + (-14.8pt,-1*\lvl)$){};
  \node (aab01) [nnf2and] at ($(oneq) + (15.2pt,-1*\lvl)$){};
  
  \node (tc1) [nnfterm] at ($(a1) + (-30pt,-.8*\lvl)$){$A$};
  \node (tc0) [nnfterm] at ($(a2) + (30pt,-.8*\lvl)$){$\neg A$};
  
  \node (te1) [nnfterm] at ($(rootLL) + (-82.5pt,-6.1*\lvl)$){$B$};
  \node (te0) [nnfterm] at ($(rootLL) + (-32.4pt,-6.1*\lvl)$){$\neg B$};
  
  \node (ta1) [nnfterm] at ($(rootLL) + (2.5pt,-7*\lvl)$){$C$};
  \node (ta0) [nnfterm] at ($(rootLL) + (32.1pt,-7*\lvl)$){$\neg C$};
  
  \node (tb1) [nnfterm] at ($(rootLL) + (107.8pt,-7*\lvl)$){$D$};
  \node (tb0) [nnfterm] at ($(rootLL) + (77.8pt,-7*\lvl)$){$\neg D$};
  
  \begin{scope}[on background layer]
  
    \draw [nnfedge] (output) -- (root.output);
    
    \draw [nnfedge] (rootLL.output) -- (root.input 2)
                        node[pos=0.4,right]  {$\ln\frac{0.6}{0.4}$};
    
    \draw [nnfedge] (a1.output) -- ++(up:0.15) -| (rootLL.input 1) 
                        node[pos=0.4,above left]  {$\ln\frac{0.9}{0.4}$};
    \draw [nnfedge] (a2.output) -- ++(up:0.15) -| (rootLL.input 2)
                        node[pos=0.4,above right]  {$\ln\frac{0.1}{0.6}$};
    \draw [nnfedge] (o12.output) -- ++(up:0.15) -| (a1.input 2);
    \draw [nnfedge] (o21.output) -- ++(up:0.15) -| (a2.input 1);
    \draw [nnfedge] (aeq1.output) -- ++(up:0.15) -| (o12.input 1)
                        node[pos=0.3,above left]  {$\ln\frac{0.2}{0.2}$};
    \draw [nnfedge] (aneq.output) -- ++(up:0.15) -| (o12.input 2)
                        node[pos=0.4,above right]  {$\ln\frac{0.8}{0.8}$};
    \draw [nnfedge] (aeq2.output) -- ++(up:0.15) -| (o21.input 2)
                        node[pos=0.3,above right]  {$\ln\frac{0.4}{0.3}$};
    \draw [nnfedge] (aneq.output) -- ++(up:0.15) -| (o21.input 1)
                        node[pos=0.4,above left]  {$\ln\frac{0.6}{0.7}$};
    \draw [nnfedge] (oeq.output) -- ++(up:0.15) -| (aeq1.input 2);
    \draw [nnfedge] (oeq.output) -- ++(up:0.15) -| (aeq2.input 2);
    \draw [nnfedge] (oneq.output) -- ++(up:0.65) -| (aneq.input 2);
    
    \draw [nnfedge] (se1.output) -- ++(up:0.27) -| (aeq1.input 1);
    \draw [nnfedge] (se2.output) -- ++(up:0.52) -| (aeq2.input 1);
    \draw [nnfedge] (se2.output) -- ++(up:0.52) -| (aneq.input 1);
    
    \draw [nnfedge] (aab11.output) -- ++(up:0.15) -| (oeq.input 1) 
                        node[pos=0.3,above left]  {$\ln\frac{0.1}{0.8}$};
    \draw [nnfedge] (aab00.output) -- ++(up:0.15) -| (oeq.input 2) 
                        node[pos=0.3,above right]  {$\ln\frac{0.9}{0.2}$};
    \draw [nnfedge] (aab10.output) -- ++(up:0.15) -| (oneq.input 1)
                        node[pos=0.3,above left]  {$\ln\frac{0.3}{0.5}$};
    \draw [nnfedge] (aab01.output) -- ++(up:0.15) -| (oneq.input 2)
                        node[pos=0.3,above right]  {$\ln\frac{0.7}{0.5}$};
    
    \draw [nnfedge] (tc1.north) -- ++(up:0.15) -| (a1.input 1);
    \draw [nnfedge] (tc0.north) -- ++(up:0.15) -| (a2.input 2);
    \draw [nnfedge] (te1.north) -- ++(up:0.15) -| (se1.input 1)
                        node[pos=0.60,above left]  {$\ln\frac{0.1}{0.6}$};
    \draw [nnfedge] (te0.north) -- ++(up:0.40) -| (se1.input 2)
                        node[pos=0.45,above right]  {$\ln\frac{0.9}{0.4}$};
    \draw [nnfedge] (te1.north) -- ++(up:0.15) -| (se2.input 1)
                        node[pos=0.25,below]  {$\ln\frac{0.8}{0.9}$};
    \draw [nnfedge] (te0.north) -- ++(up:0.40) -| (se2.input 2)
                        node[pos=0.47,above right]  {$\ln\frac{0.2}{0.1}$};
    
    \draw [nnfedge] (ta1.north) -- ++(up:0.15) -| (aab11.input 1);
    \draw [nnfedge] (ta1.north) -- ++(up:0.15) -| (aab10.input 1);
    \draw [nnfedge] (ta0.north) -- ++(up:0.35) -| (aab01.input 1);
    \draw [nnfedge] (ta0.north) -- ++(up:0.35) -| (aab00.input 1);
    
    \draw [nnfedge] (tb1.north) -- ++(up:0.55) -| (aab11.input 2);
    \draw [nnfedge] (tb1.north) -- ++(up:0.55) -| (aab01.input 2);
    \draw [nnfedge] (tb0.north) -- ++(up:0.75) -| (aab00.input 2);
    \draw [nnfedge] (tb0.north) -- ++(up:0.75) -| (aab10.input 2);
    
  \end{scope}
\end{tikzpicture}
	}
    	\caption{Logistic circuit for $\Pr(Y=1 \mid A,B,C,D)$} \label{circuit: prop circuit: equivtoprob}
	\end{subfigure}
\caption{A probabilistic circuit with parallel structures under class variable $Y$ and its equivalent logistic circuit for predicting~$Y$}
\label{circuit: prop circuit}
\end{figure*}

\subsection{Model Complexity \& Data Efficiency}
Table~\ref{table: size} summarizes the size of all compared models when achieving the reported  accuracy. We can conclude that logistic circuits are significantly smaller than the alternatives, despite attaining higher accuracy. 

We design the next set of experiments to specifically investigate how well our structure learning algorithm performs under the setting where the number of training samples is limited.  We have two additional sets of experiments, where only $2\%$ and $10\%$ of the original training data is supplied. 
Table~\ref{table: data efficiency} summarizes the performance in this limited-data setting. 
We mainly compare against a 5-layer MLP and CNN with 3 convolutional layers, whose performance is on par with our method under the full-data setting. As summarized in Table~\ref{table: data efficiency}, except on MNIST with $10\%$ training samples, real-valued logistic circuits achieve the best classification accuracy. Moreover, in both versions of logistic circuits, when the available training samples are reduced from $100\%$ to $2\%$, the accuracy only drops by around $3\%$ when evaluating on MNIST; around $5\%$ on Fashion. In contrast, a much larger drop occurs for 5-layer MLP and CNN. Specifically, MLP's accuracy drops by $5\%$ ($9\%$) while CNN's accuracy drops by $4\%$ ($7\%$) on MNIST (Fashion). This small magnitude of accuracy decrease illustrates how data efficient our proposed structure learning algorithm is.

Except on MNIST with $10\%$ training samples, real-valued logistic circuits achieve the best classification accuracy. From a top-down perspective, each OR gate of a logistic circuit presents a weighted choice between its wires. Hence, one can view a logistic circuit as a decision diagram.
Under this perspective, splits refine OR gates' branching rules. As each branching rule naturally applies to multiple samples, we hypothesize that the splits selected by our structure learning algorithm reflect the general conditional feature information present in the dataset.

\subsection{Local Explanation}
\label{s: interpretability}
Next, we aim to share some insights about how to explain the learned logistic circuit. Specifically, we investigate the question: ``Why does the logistic circuit classify a given sample $\sample$ as $y$?'' Since any logistic circuit can be reduced to a logistic regression classifier, we can easily find the active global flow feature that contributes most to the given sample's classification probability. That is, the feature that maximizes $\mathbbm{x}\cdot\theta$. We visualize one such feature for MNIST data and one for Fashion in Figure~\ref{fig: feature visualization} by marking the variables used in the their corresponding logical sentences.

\section{Connection to Probabilistic Circuits} \label{s:generativeconnection}
In recent years, a large number of tractable probabilistic models have been proposed as a target representation for generative learning of a joint probability distribution: arithmetic circuits~\cite{lowd:uai08}, weighted SDD~\cite{BekkerNIPS15}, PSDD \cite{KisaVCD14}, cutset networks~\cite{rahman2014cutset} and sum-product networks (SPNs) \cite{poon2011sum}.
These representations have various syntactic properties. Some put probabilities on terminals, others on edges. Some use logical notation (AND, OR), others use arithmetic notation ($\times$,$+$).
Nevertheless, they are all circuit languages built around the properties of decomposability and/or determinism.

For our purpose, we consider a simple probabilistic circuit language based on the logistic circuit syntax, where now the $\theta$ parameters are assumed to be  normalized probabilities.\footnote{We also assume \emph{smoothness}~\cite{darwicheJAIR02}.}
\begin{definition}[Probabilistic Circuit Semantics]
\label{de: probabilistic circuit semantics}
A probabilistic circuit node $n$ defines the following joint distribution.
\begin{itemize}
\item[--] If $\node$ is a leaf (input) node, then $\Pr_n(\sample) =[\sample \models n]$.

\item[--] If $\node$ is an AND gate with children $c_1,\dots,c_m$, then
\begin{align*}
\Pr{_n}(\sample) =  \prod_{i=1}^m \Pr{_{c_i}}(\sample).
\end{align*}

\item[--] If $\node$ is an OR gate with (child node, wire parameter) inputs $(c_1,\theta_1),\dots, (c_m, \theta_m)$, then
\begin{align*}
\Pr{_n}({\bf x}) =    \sum_{i=1}^m  \Pr{_{c_i}}(\sample)\cdot \theta_i.
\end{align*}
\end{itemize}
\end{definition}

Figure~\ref{circuit: prop circuit: prob } shows a probabilistic circuit for the joint distribution $\Pr(Y,A,B,C,D)$. This tractable circuit language is a relaxation of PSDDs \cite{KisaVCD14} and a specific type of SPN \cite{poon2011sum} where determinism holds throughout. It is also a type of arithmetic circuit.

We are now ready to connect logistic and probabilistic circuits. It is well known that logistic regression is the discriminative counterpart of a naive Bayes generative model~\cite{ng2002discriminative}. A similar correspondence holds between our logistic and probabilistic circuits.
\begin{proposition}
\label{prop: correspondence}
Consider a probabilistic circuit whose structure is of the form $(Y \land \alpha) \lor (\neg Y \land \beta)$, where sub-circuits $\alpha$ and $\beta$ are structurally identical.
Then, there exists an equivalent logistic circuit for the conditional probability of $Y$ in the probabilistic circuit. Moreover, this logistic circuit has structure $\lor \alpha$ and its parameters can be computed in closed form as log-ratios of probabilistic circuit probabilities.
\end{proposition}

We first depict this correspondence intuitively in Figure~\ref{circuit: prop circuit}. The logistic circuit has the same structure as the two halves of the probabilistic circuit, and its parameters are computed from the probabilistic circuit probabilities. The distributions $\Pr(Y=1 \mid A,B,C,D)$ represented by the circuits in Figures~\ref{circuit: prop circuit: prob } and~\ref{circuit: prop circuit: equivtoprob} are identical. 

\paragraph{Formal Correspondence}
Next, we present the formal proof of this correspondence for binary $\sample$. 
Recall that in our circuits, only the input wires of OR gates are parameterized. Let $\mathcal{W}_\delta$ be the set that contains all these wires in circuit~$\delta$: $$\mathcal{W}_\delta = \left\{(n, c) \mid c\text{ is a gate with parent OR gate } n \right\}.$$ 
After expanding the equations in Definition~\ref{de: probabilistic circuit semantics} and following the top-down definition of global circuit flow (i.e., following Definition~\ref{definition: global flow}), one finds that the joint distribution induced by a probabilistic circuit $\delta$ can be rewritten as
$$
\Pr{_\delta}(\sample) = \prod_{(n, c) \in \mathcal{W}_\delta} f_\delta(n,\sample,c) \cdot \theta_{(n,c)}^\delta.
$$
We will exploit this finding in the derivation of the conditional distribution induced by the probabilistic circuit~$\gamma = (Y \land \alpha) \lor (\neg Y \land \beta)$.
\begin{align}
&\Pr{_{\gamma}}(Y=1 \mid \sample) \nonumber \\
 &\quad =\frac{\Pr_\gamma(Y\!=\!1)\Pr_\alpha(\sample)}{\Pr_\gamma(Y\!=\!0)\Pr_\beta(\sample) + \Pr(Y\!=\!1)\Pr_\alpha(\sample)} \nonumber\\
&\quad = \frac{1}{1+\frac{\Pr_{\gamma}(Y=0)\Pr_\beta(\sample)}{\Pr_{\gamma}(Y=1)\Pr_\alpha(\sample)}} \nonumber \\
&\quad =\frac{1}{1+\frac{\Pr_{\gamma}(Y=0)\prod_{(n,c) \in \mathcal{W}_\beta} f_\beta(n,\sample,c)\theta_{(n,c)}^\beta}{\Pr_{\gamma}(Y=1)\prod_{(n,c) \in \mathcal{W}_\alpha} f_\alpha(n,\sample,c)\theta^\alpha_{(n,c)}}} \nonumber
\end{align}
As stated in Proposition~\ref{prop: correspondence} and shown in Figure~\ref{circuit: prop circuit}, sub-circuits $\alpha$ and $\beta$ share the same structure. Therefore, we can further simplify this equation as follows.
\begin{align}
&\Pr{_{\gamma}}(Y=1 \mid \sample) \nonumber \\
&\quad = \frac{1}{1+\frac{\Pr_{\gamma}(Y=0)}{\Pr_{\gamma}(Y=1)}\prod_{(n,c) \in \mathcal{W}_{\alpha} } f_{\lor \alpha}(n, \sample, c) \frac{\theta^\beta_{(n,c)}}{\theta^\alpha_{(n,c)}} }  \nonumber
 \\
&\quad = \frac{1}{1+\exp\left[- g(\sample)) \right]} = \Pr{_{\lor \alpha}}(Y=1 \mid \sample) \nonumber
\end{align}
where
\begin{align}
g(\sample)&= \log\frac{\Pr_\gamma(Y\!=\!1)}{\Pr_\gamma(Y\!=\!0)} + \!\!\!\sum_{(n,c) \in \mathcal{W}_{\alpha}} \!\!  f_{\lor \alpha}(n,\sample,c) \log \frac{\theta^\alpha_{(n,c)}}{\theta^\beta_{(n,c)}}  \label{eq:nb}\\
&=\theta^{\lor \alpha}_{\mathit{root}} + \sum_{(n,c) \in \mathcal{W}_{\alpha}}f_{\lor \alpha}(n,\sample,c) \cdot \theta_{(n,c)}^{\lor \alpha}. \label{eq:lr}
\end{align}
The transformation from Equation~\ref{eq:nb} to~\ref{eq:lr} expresses the logistic circuit parameters as the  log-ratios of probabilistic circuit probabilities. 
For example, the class priors captured in the output wires of $\alpha$ and $\beta$ are now combined as a log-ratio to form the bias term for $\lor \alpha$, expressed by the root parameter.

This proof also provides us with a new perspective to understand the semantics of the learned parameters.  
The parameters represent the log-odds ratio of the features given different classes. Note that by Bayes' theorem, a naive Bayes model would derive its induced distribution in a sequence of steps similar to the ones above, resulting in Equation~\ref{eq:nb}. Given this correspondence, one can also view our proposed structure learning method as a way to construct meaningful features for a naive Bayes classifier. We know that after training, naive Bayes classifiers are equivalent to logistic regression classifiers (as in Equation~\ref{eq:lr}).

 \section{Related Work}
\citet{gens2012discriminative} proposed the first parameter learning algorithm for discriminative SPNs, using MPE inference as a sub-routine. Without the support of the determinism property, parameter learning of general SPNs is a relatively harder question than its logistic circuit counterpart, since it is non-convex. \citet{adel2015learning} boost the accuracy of SPNs on MNIST to  $97.6\%$ by extracting more representative features from raw inputs based on the Hilbert-Schmidt independence measure. 
\citet{rat-spn2018} further improved the  classification ability of SPNs by drastically simplifying SPN structure requirements and utilizing a loss objective that hybrids cross-entropy (discriminative learning) with log-likelihood (generative learning).

\citet{rooshenas2016discriminative} developed a discriminative structure learning algorithm for arithmetic circuits. The method updates the circuit that represents a corresponding conditional random field (CRF) model by adding features conditioned on arbitrary evidence to the model. This work further relaxes decomposability and smoothness properties of ACs for a more compact representation. However, it targets the setting where there are a large number of output variables, not single-variable classification.

All the aforementioned literature conforms to a common trend of abandoning properties of the chosen circuit representations for easier structure learning and better prediction accuracy. They argue that those special syntactic restrictions complicate the learning process. On the contrary, this paper chooses perhaps the most structure-restrictive circuit as the target representation. Instead of relaxing the target representation's syntactical requirements, our proposed method fully leverages the valuable properties that stem from these restrictions, and in particular convexity. 

\section{Conclusions}

We have presented logistic circuits, a novel circuit-based classification model with convex parameter learning and a simple structure learning procedure based on local search. Logistic circuits outperform much larger classifiers and perform well in a limited data regime. Compared to other symbolic, circuit-based approaches, logistic circuits present a leap in performance on image classification benchmarks.
Future work includes support for convolution, parameter tying, and structure sharing in the logistic circuits framework.

\appendix 
  
   \begin{algorithm}[t]
     \caption{Node probabilities from a real-valued sample $\sample$.}
     \label{alg: node probability}
      \LinesNumbered         
      \SetKwInOut{Input}{Input}
      \SetKw{KwAnd}{and}
      \SetKw{KwOr}{or}
      \DontPrintSemicolon
      \Input{A vector of probabilities $\sample$.  }
      \KwResult{$\Pr_\sample(\node)$: the node probability of $\node$ for $\sample$.}
      \BlankLine 
      \For{$\node$ in the circuit's nodes, children before parents}
      { \If{\node~\text{is a leaf with variable $X$}~}{
      	\If{$\node$ is $X$} {$\Pr_\sample(\node) = \sample(X)$}
	\Else{
	%\tcp{$\node$ is $\neg X$}
	$\Pr_\sample(\node) = 1 - \sample(X)$
	}
      }
      \ElseIf{\node~\text{is an AND gate}
      }{
      $\Pr_\sample(\node) := 1$\\
      \For{$c$ in inputs of $\node$}{
      $\Pr_\sample(\node)~*= \Pr_\sample(c)$
       }
      }\Else{
      \tcp{$\node$ is an OR gate}
      $\Pr_\sample(\node) := 0$\\
      \For{$c$ in inputs of $\node$}{
      $\Pr_\sample(\node)~+= \Pr_\sample(c)$
       }
      }
      }
  \end{algorithm} 
  
          \begin{algorithm}[t]
                \caption{Features $\bm{\mathbbm{x}}$ from a real-valued sample $\sample$.}
                       \label{alg: real-valued feature calculation}
	\LinesNumbered
	\SetKwInOut{Input}{Input}
      \SetKw{KwAnd}{and}
      \SetKw{KwOr}{or}
      \DontPrintSemicolon
      \Input{Node probabilities $\Pr_\sample(\cdot)$.}
      \KwResult{Real-valued feature vector $\bm{\mathbbm{x}}$.}
      \BlankLine
        \For{$\node$ in all nodes, parents before children}{
        $v(\node) := 0$
        }

    $v(\text{root}) := 1$\\
   \For{$\node$ in all non-leaf nodes, parents before children}{
   \If{$\node$~\text{is an OR gate}}{
    \For{$c$ in inputs of $\node$}{
      $\bm{\mathbbm{x}}(n, c) := v(\node) \cdot \Pr_\sample(c)~/~\Pr_\sample(\node)$ \\
      $v(c)~+= \bm{\mathbbm{x}}(\node, c)$
	}
    }
    \Else{
    \tcp{$\node$ is an AND gate}
    \For{$c$ in inputs of $\node$}{
    $v(c)~+= v(n)$	
     }
    }
     }
     
         \end{algorithm} 
 
 \section{Proof of Proposition~\ref{proposition: features}}
\label{section: proof of proposition}
 Before presenting the proof, we restate the proposition.
 \begin{proposition*}
The features $\mathbbm{x}$ constructed in the proof of Proposition~\ref{proposition: logistic regression} are equivalent to global flows $f_r(n,\sample,c)$.
 \end{proposition*}
In the following, we prove this proposition by induction.
 \begin{itemize}
\item[--] \underline{Base case}: the inputs of the root $r$ are either leaf nodes or AND gates whose inputs are leaf nodes. By definition, for the root's input wires, their local circuit flow equals their global circuit flow. According to the decomposition matrix of $g_n$ in the proof of Proposition~\ref{proposition: logistic regression}, the features associated with the root's input wires are equivalent to their local circuit flow. By transitivity, we prove logistic circuits' features are equivalent to its global circuit flow vector in the base case.
\item[--] \underline{Induction step}: assume the proposition holds for all OR gates in a given logistic circuit except the root $r$. Again, the root's inputs can be either leaf nodes or AND gates. It is obvious that for the root's input wires, their associated features are equivalent to their global circuit flow, as this has been proven in the base case.
So we only need to focus on the wires of the sub logistic circuits rooted on those AND gates. The inputs to those AND gates can either be leaf nodes or OR gates. As the wires between AND gates and their leaf children do not have parameters, the correctness of the proposition does not get affected by them. We can narrow our focus again. Now let us consider an OR gate $n$, which is an input to some of those aforementioned AND gates $\{e_1,\dots, e_m\}$. By our induction assumption, its features are equivalent to the global circuit flows defined with respect to $n$; in other words, $\bm{\mathbbm{x}}_n = f_n$. After propagating $\bm{\mathbbm{x}}_n$ upwards to the root, we get $\sum_{i=1}^m f(r, {\bf x}, e_1)\cdot \bm{\mathbbm{x}}_n$. The sum of the global flow on all output wires of $n$ is $F_r(n) = \sum_{i=1}^m f(r, {\bf x}, e_1) $. Since $F_r(n)$ is propagated throughout the whole sub logistic circuit rooted at $n$, the global circuit flow in this sub logistic circuit with respect to the root $r$ is $F_r(n)\cdot f_n =  \sum_{i=1}^m f(r, {\bf x}, e_1) \cdot f_n$. Therefore, the constructed features are equivalent to the global circuit flows.
\end{itemize}

 \section{Calculation of Node Probabilities}
 \label{section: node probabilities}
We calculate node probabilities in a bottom-up induction on the structure of the sentence.

\begin{itemize}
\item[--] Base case: $n$ is a leaf (input) node. The node probability is directly defined in $\sample$: $\Pr_\sample(n) = \sample(X)$ if $n$ is $X$; $\Pr_\sample(n) = 1 - \sample(X)$ if $n$ is $\neg X$ (lines 2-6 in Algorithm~\ref{alg: node probability}).
\item[--] Induction step: given that the node probabilities for all the leaves have been calculated, we move upward to intermediate nodes and the root, where there are two cases.
\begin{itemize}
 \item[*] $n$ is an AND gate with inputs $\left\{c_1,\dots,c_m \right\}$. Since in a logistic circuit every AND gate is decomposable, by independence of the conjuncts, $\Pr_\sample(n) = \prod_{i=1}^m \Pr_\sample(c_i)$ (lines 7-10 in Algorithm~\ref{alg: node probability}).
 \item[*] $n$ is an OR gate with input nodes $\left\{c_1,\dots,c_m\right\}$. Since every OR gate is deterministic, the probabilistic events defined at each child within the same OR parent do not intersect with each other. By mutual exclusivity, $\Pr_\sample(n) = \sum _i \Pr_\sample(c_i)$ (lines 11-14 in Algorithm~\ref{alg: node probability}).
 \end{itemize}
\end{itemize}
        
 \section{Calculation of Global Flows (Features)}
 \label{appendix: feature calculation}
 Node probabilities $\Pr_\sample(\cdot)$ are used in Algorithm~\ref{alg: real-valued feature calculation} to obtain the final feature vector. 

 We perform a top-down pass starting from the root OR gate. After visiting an OR gate,  the method first calculates its associated global circuit flows from its inputs; see Line~7 in Algorithm~\ref{alg: real-valued feature calculation}. These newly calculated global flows then get passed down and are accumulated on those child gates for later use on the descendent gates (Line 8). After visiting an AND gate, there is no new global circuit flow to be calculated. Hence, the algorithm directly accumulates the flows passed to those AND gates to their children (Line 10-11).
 \label{s: calculation of global flows}

Note that instead of inputing one single sample at a time, one can directly supply Algorithm~\ref{alg: node probability} and \ref{alg: real-valued feature calculation} with a vector of samples. Our proposed calculation method is completely compatible with matrix operations, and by doing so, one can expect a large speedup.
	
  \section{Initial Structure}
  \label{appendix: initial structure}
  \begin{figure}[t!]
	\centering
     		\scalebox{0.72}{
		\begin{tikzpicture}[circuit logic US, nnf]
	
	\def\lvl{25pt}        
	          
	\node (at) [nnfterm] at (0,0) {$~~A~$};
	\node (bt) [nnfterm] at ($(at) + (\lvl,0)$) {$~~B~$};
	\node (af) [nnfterm] at  ($(at) + (2*\lvl,0)$){$\neg A$};
	\node (bf) [nnfterm] at ($(at) + (3*\lvl,0)$) {$\neg B$};
	
	\node (ct) [nnfterm] at  ($(at) + (5*\lvl,0)$){$~~C~$};
	\node (dt) [nnfterm] at ($(at) + (6*\lvl,0)$) {$~~D~$};
	\node (cf) [nnfterm] at  ($(at) + (7*\lvl,0)$){$\neg C$};
	\node (df) [nnfterm] at ($(at) + (8*\lvl,0)$) {$\neg D$};
	
	\node (and1) [nnfand, inputs=nn, scale=0.85] at ($(at) + (0.09*\lvl,1.5*\lvl)$) {};
	\node (and2) [nnfand, inputs=nn, scale=0.85] at ($(bt) + (0.09*\lvl,1.5*\lvl)$) {};
	\node (and3) [nnfand, inputs=nn, scale=0.85] at ($(af) + (0.09*\lvl,1.5*\lvl)$) {};
	\node (and4) [nnfand, inputs=nn, scale=0.85] at ($(bf) + (-0.09*\lvl,1.5*\lvl)$) {};
	
	\draw[nnfedge] (at) --++ (up:0.45) -| (and1.input 1);
	\draw[nnfedge] (bt) --++ (up:0.45) -| (and1.input 2);
	
	\draw[nnfedge] (at) --++ (up:0.75) -| (and2.input 1);
	\draw[nnfedge] (bf) --++ (up:0.75) -| (and2.input 2);
	
	\draw[nnfedge] (af) --++ (up:0.45) -| (and3.input 1);
	\draw[nnfedge] (bt) --++ (up:0.45) -| (and3.input 2);
	
	\draw[nnfedge] (af) --++ (up:0.6) -| (and4.input 1);
	\draw[nnfedge] (bf) --++ (up:0.6) -| (and4.input 2);

	\node (and5) [nnfand, inputs=nn, scale=0.85] at ($(ct) + (0.09*\lvl,1.5*\lvl)$) {};
	\node (and6) [nnfand, inputs=nn, scale=0.85] at ($(dt) + (0.09*\lvl,1.5*\lvl)$) {};
	\node (and7) [nnfand, inputs=nn, scale=0.85] at ($(cf) + (0.09*\lvl,1.5*\lvl)$) {};
	\node (and8) [nnfand, inputs=nn, scale=0.85] at ($(df) + (-0.09*\lvl,1.5*\lvl)$) {};
	
	\draw[nnfedge] (ct) --++ (up:0.45) -| (and5.input 1);
	\draw[nnfedge] (dt) --++ (up:0.45) -| (and5.input 2);
	
	\draw[nnfedge] (ct) --++ (up:0.75) -| (and6.input 1);
	\draw[nnfedge] (df) --++ (up:0.75) -| (and6.input 2);
	
	\draw[nnfedge] (cf) --++ (up:0.45) -| (and7.input 1);
	\draw[nnfedge] (dt) --++ (up:0.45) -| (and7.input 2);
	
	\draw[nnfedge] (cf) --++ (up:0.6) -| (and8.input 1);
	\draw[nnfedge] (df) --++ (up:0.6) -| (and8.input 2);

	\node (or1) [nnfor, inputs=nnnn, scale=0.65] at ($(and2) + (0.5*\lvl, 1.5*\lvl)$) {};
	
	\draw[nnfedge] (and1) --++ (up:0.75) -| (or1.input 1);
	\draw[nnfedge] (and2) --++ (up:0.6) -| (or1.input 2);
	\draw[nnfedge] (and3) --++ (up:0.6) -| (or1.input 3);
	\draw[nnfedge] (and4) --++ (up:0.75) -| (or1.input 4);

	\node (or2) [nnfor, inputs=nnnn, scale=0.65] at ($(and6) + (0.5*\lvl, 1.5*\lvl)$) {};
	
	\draw[nnfedge] (and5) --++ (up:0.75) -| (or2.input 1);
	\draw[nnfedge] (and6) --++ (up:0.6) -| (or2.input 2);
	\draw[nnfedge] (and7) --++ (up:0.6) -| (or2.input 3);
	\draw[nnfedge] (and8) --++ (up:0.75) -| (or2.input 4);

	\node (and9) [nnfand, inputs=nn, scale=1.05] at ($(or1) + (2.5*\lvl, 1.5*\lvl)$) {};
	
	\draw[nnfedge] (or1) -- ++ (up:0.6) -| (and9.input 1);
	\draw[nnfedge] (or2) --++ (up:0.6) -| (and9.input 2);
	
	\node (root) [nnfor, inputs=nnn, scale=0.85] at ($(and9) + (0, 1.2*\lvl)$) {};
	
	\draw[nnfedge] (and9) -- ++ (up:0.6) -| (root.input 2);
              
        \end{tikzpicture}
}
		\caption{Initial structure of logistic circuits with 4 pixels.}\label{fig: initial structure}
\end{figure}
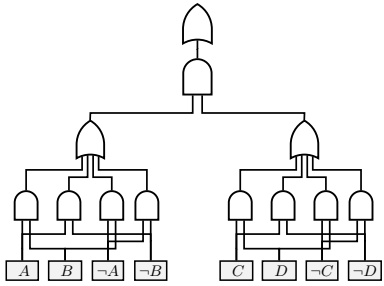
  All experiments in this paper start with an initial structure where every pixel has two corresponding leaf nodes, one for the pixel being true and the other false. 
  Pixels are paired up by AND gates; an AND gate is created for every joint assignment to the pair.
   AND gates for the same pair share one OR gate parent. After this, OR gates are paired with AND gates and every AND gate is connected to its own OR gate parent until we reach the root. Figure~\ref{fig: initial structure} is an example of the initial structure with 4 pixels. Note that our structure learning algorithm is compatible with other initial structures and one can create ad-hoc ones tailored to different applications.
  
\section{Details of Existing Classification Models}
\label{s: model details}
   The reported kernel logistic regression is based on the pixel n-grams implemented in Vowpal Wabbit \cite{vw2007}. The reported random forest has 500 decision trees. The reported SVM with RBF Kernel uses hyper-parameters $C=8, \gamma=0.05$ on MNIST and $C=4,\gamma=25$ on Fashion. The reported 3-layer MLP has layers of size 784-1000-500-250-10 respectively. The reported 5-layer MLP has layers of size 784-1000-500-250-2000-250-10 respectively. The reported CNN with 3 convolutional layers uses 3-by-3 padded filters in the convolutional layers.

\section*{Acknowledgements}
This work is partially
supported by
a gift from Intel,
NSF grants \#IIS-1657613, \#IIS-1633857, \#CCF-1837129,
and DARPA XAI grant \#N66001-17-2-4032.
\bibliographystyle{aaai}
\bibliography{reference}
\end{document}